\documentclass[conference]{IEEEtran}
\IEEEoverridecommandlockouts

\usepackage{cite}
\usepackage{amsmath,amssymb,amsfonts}
\usepackage{algorithm}
\usepackage{algpseudocode}
\usepackage{nicefrac}
\usepackage{graphicx}
\usepackage{textcomp}
\usepackage{xcolor}
\usepackage{xspace}
\usepackage{amsthm}
\usepackage{multirow}
\usepackage{adjustbox}
\usepackage{enumitem}
\usepackage{makecell}
\usepackage{xurl}
\usepackage{thm-restate}
\usepackage{comment}
\usepackage{subfig}
\usepackage{cleveref}
\usepackage{caption}
\allowdisplaybreaks

\def\BibTeX{{\rm B\kern-.05em{\sc i\kern-.025em b}\kern-.08em
    T\kern-.1667em\lower.7ex\hbox{E}\kern-.125emX}}

\newcommand{\CBayes}{\mathcal{C}_{\mathrm{Bayes}}}
\newcommand{\Bayes}{\ensuremath{\mathrm{bv}}}
\newcommand{\gbar}{\overline{\mathbf{g}}}
\newcommand{\gtilde}{\tilde{\mathbf{g}}}

\newcommand{\VM}{\mathcal{V}}
\newcommand{\Gauss}{\mathcal{N}}
\newcommand{\Alg}[1]{Algorithm~\ref{#1}}
\newcommand{\Eqn}[1]{Eqn (\ref{#1})}

\newcommand{\Def}[1]	{Definition \ref{#1}}

\newcommand{\Lem}[1]    {Lemma \ref{#1}}
\newcommand{\Sec}[1]	{Section \ref{#1}}

\newcommand{\DirDP}{\textsc{DirDP-SGD}\xspace}

\newcommand{\IGA}{IGA\xspace}
\newcommand{\Reals}{\mathbb{R}_{\geq 0}}
\newcommand{\calw}  {\ensuremath{\mathcal{W}}}
\newcommand{\calx}  {\ensuremath{\mathcal{X}}}
\newcommand{\caly}  {\ensuremath{\mathcal{Y}}}
\newcommand{\calz}  {\ensuremath{\mathcal{Z}}}
\newcommand{\Dist}  {\ensuremath{\mathbb{D}}}
\newcommand{\supp}[1] {\ensuremath{\lceil{#1}\rceil}}

\newcommand {\citet} {\cite}
\newcommand {\citep} {\cite}

\newcommand{\VPrior}        {\ensuremath{V_g}}
\newcommand{\VgPrior}[1]    {\ensuremath{V_{#1}}}
\newcommand{\VPosterior}    {\ensuremath{V_g}}
\newcommand{\VgPosterior}[1]   {\ensuremath{V_{#1}}}

\newcommand{\Leak} 	{\mathcal{L}_g}
\newcommand{\gLeak}[1] 	{\mathcal{L}_{#1}}

\definecolor{greenmunsell}{rgb}{0.0, 0.66, 0.47}

\newcommand{\SBnote}[2]{{\color{brown}SB note---#1: #2}}

\newcommand{\CPchange}[1]{#1}

\newcommand{\tashsays}[1]{{\color{teal}#1}}

\theoremstyle{definition}
\newtheorem{definition}{Definition}
\theoremstyle{remark}

\newtheorem{theorem}{Theorem}
\newtheorem{corollary}{Corollary}
\newtheorem{proposition}{Proposition}
\newtheorem{lemma}{Lemma}

\begin{document}

\title{Comparing privacy notions for protection against reconstruction attacks in machine learning 
}


\author{
\IEEEauthorblockN{Sayan Biswas}
\IEEEauthorblockA{
\textit{EPFL}\\
Lausanne, Switzerland \\
sayan.biswas@epfl.ch}
\and
\IEEEauthorblockN{Mark Dras}
\IEEEauthorblockA{
\textit{Macquarie University}\\
Sydney, Australia \\
mark.dras@mq.edu.au}
\and
\IEEEauthorblockN{Pedro Faustini}
\IEEEauthorblockA{
\textit{Macquarie University}\\
Sydney, Australia \\
pedro.arrudafaustini@hdr.mq.edu.au}
\and
\IEEEauthorblockN{Natasha Fernandes}
\IEEEauthorblockA{
\textit{Macquarie University}\\
Sydney, Australia \\
natasha.fernandes@mq.edu.au}
\and
\IEEEauthorblockN{Annabelle McIver}
\IEEEauthorblockA{
\textit{Macquarie University}\\
Sydney, Australia \\
annabelle.mciver@mq.edu.au}
\and
\IEEEauthorblockN{Catuscia Palamidessi}
\IEEEauthorblockA{
\textit{INRIA and \'Ecole Polytechnique}\\
Palaiseau, France \\
catuscia@lix.polytechnique.fr}
\and
\IEEEauthorblockN{Parastoo Sadeghi}
\IEEEauthorblockA{
\textit{UNSW Canberra}\\
Australia \\
p.sadeghi@unsw.edu.au}
}
   


\maketitle
\thispagestyle{empty}
\pagestyle{empty}

\begin{abstract}
    Within the machine learning community, reconstruction attacks are a principal concern and have been identified even in federated learning (FL), which was designed with privacy preservation in mind. In response to these threats, the privacy community recommends the use of differential privacy (DP) in the stochastic gradient descent algorithm, termed DP-SGD. However, the proliferation of variants of DP in recent years—such as metric privacy—has made it challenging to conduct a fair comparison between different mechanisms due to the different meanings of the privacy parameters $\epsilon$ and $\delta$ across different variants. Thus, interpreting the practical implications of \(\epsilon\) and \(\delta\) in the FL context and amongst variants of DP remains ambiguous. In this paper, we lay a foundational framework for comparing mechanisms with differing privacy guarantees, namely \((\epsilon, \delta)\)-privacy and metric privacy. We provide two foundational means of comparison: firstly, via the well-established \((\epsilon, \delta)\)-DP guarantees, made possible through the Rényi differential privacy framework; and secondly, via Bayes' capacity, which we identify as an appropriate measure for reconstruction threats.
\end{abstract}


\begin{IEEEkeywords}
R\'enyi differential privacy, metric privacy, Bayes capacity, reconstruction attacks, stochastic gradient descent
\end{IEEEkeywords}

\section{Introduction}\label{sec:intro}

\CPchange{In the context of machine learning (ML) and federated learning (FL), privacy concerns have been addressed mainly by \emph{differential privacy} (DP)~\citep{dwork-roth:2014} and its variants, such as approximate DP and \emph{metric privacy} (aka $d_{\mathcal{X}}$-privacy)~\citep{chatzikokolakis-etal:2013:PETS}. Although all these definitions use a parameter $\epsilon$ to represent the level of privacy, the meaning of $\epsilon$ in metric privacy is different from that in DP, and this has generated quite some confusion. For instance,  metric privacy may use extremely high values of $\epsilon$ (in the order of hundreds) and still offer a good protection. The main goal of this paper is to establish a framework to compare the privacy-utility trade-off between DP  and metric privacy mechanisms in scenarios where a significant source of vulnerability comes from \emph{reconstruction attacks}.


The first proposals to protect privacy in ML adopted standard DP: 
 A modification of the widely used stochastic gradient descent (SGD) to incorporate differential privacy (DP-SGD) was proposed by \cite{song-etal:2013} and further developed by \cite{abadi-etal:2016:CCS}. Until now, most of the works in privacy-protecting ML and FL have been based on DP-SGD, and have applied, most commonly, the Gaussian mechanism~\citep{dwork-roth:2014}. 
 
 In recent years, however, in the quest to find a good utility-privacy trade-off, there has been an increasing interest in using \emph{metric privacy} primarily in the domain of location-privacy~\cite{OyaMILocationPrivacy2017,biswasprivic2023,UgurPrivacyPreservingEVs2024}. This is a metric-based generalisation of DP in which the ratio of the probabilities of two inputs being mapped to the same obfuscated value is bound by 
an exponential expression that depends not only on the privacy parameter  $\epsilon$, but also on the distance between the two inputs.  In ML, metric privacy has been applied 
in the context of image processing (e.g., \cite{croft-etal:2019,croft-etal:2022:FGCS}), natural language processing (e.g.,  \cite{fernandes-etal:2019:POST,feyisetan-etal:2019,klymenko-etal-2022-differential}), personalized FL (e.g.,  \cite{galli2022group, galli2023advancing}), etc.
Most of these metric privacy-based proposals use the Laplace mechanism. One exception is \cite{faustini2023directional}, which uses the von Mises Fisher (VMF) mechanism introduced in~\citep{weggenmann-kerschbaum:2021:CCS}. While the Gaussian and Laplace mechanisms are isotropic, i.e., equally likely to perturb the gradient vector in any direction in its high-dimensional space, the application of VMF adds angular noise and preserves the norm. 

The Gaussian mechanism provides approximate DP measured in terms of two parameters $\epsilon$ and $\delta$. Metric private mechanisms also use a parameter $\epsilon$, but with a different meaning. 
Moreover, in both cases, the parameters measure the protection against \emph{membership inference attacks} (MIAs), and are not necessarily representative of the robustness with respect to other types of attacks. 
In ML, and especially in FL, \emph{reconstruction attacks} are generally considered to be one of the main threats to privacy (e.g., \cite{melis-etal:2019:IEEE-SP, balle-etal:2022:IEEE-SP}). 
These attacks consist of trying to guess the exact value of the training data after noisy gradient observations. 
The natural question that arises then is: \emph{how do we measure and compare the ability of different perturbation mechanisms to defend against reconstruction attacks?}

Some of the above works have conducted experimental comparisons between the proposed metric privacy mechanisms and the DP ones. However, experimental results are limited to the specific case under consideration, and do not allow to draw general conclusions. 
The main purpose of this paper is to provide a \emph{theoretical foundation} for measuring and comparing the privacy-utility trade-off between different methods of perturbation of the popular SGD algorithm. We focus on the privacy vulnerability of reconstruction and, as an example, we apply our measure
 to compare two very different types of perturbation used in the ML community:  Gaussian and VMF. 
 We emphasize that the purpose is not to argue that one of our chosen
mechanisms or privacy definitions is better than the other, but rather to use them to showcase a proper comparison framework
and to develop sound principles and measures for comparison based on an actual threat scenario, namely reconstruction.

We first consider  R\'enyi DP (RDP)~\cite{8049725} as a unifying framework, due to its ability to accommodate multiple DP definitions. Indeed, RDP has already been used in the literature as a common metric for $\epsilon$-DP and  $(\epsilon,\delta)$-DP mechanisms, and relations between the parameters of DP and those of RDP are well established. We show that it can also be used to measure metric private mechanisms by converting the parameters of the VMF mechanism into the parameters of RDP. The further conversion from the parameters of RDP into the $(\epsilon,\delta)$ of (approximate) DP allows to compare directly the different mechanisms on the basis of the same parameters. 

The remaining question is whether RDP and $(\epsilon, \delta)$ represent well reconstruction attacks.  It has been observed that the values of $\epsilon$   for the Gaussian and VMF mechanisms do not correlate with the protections that these mechanisms provide against reconstruction attacks~\cite{faustini2023directional}.
One could argue that the parameters considered in the above work were incomparable, due to one referring to standard DP and the other to metric privacy.  However, the new experiments in this paper show that, even when compared using a common differential privacy framework (via RDP), the values computed for $\epsilon$  and $\delta$ do not correspond to effectiveness in protecting against reconstruction attacks.
A similar observation, namely that $\epsilon$ does not account for all kinds of attacks,  was already made in \cite{Chatzi:2019}.


To address the inadequacy of  RDP and $(\epsilon,\delta)$-DP to represent the protection against reconstruction attacks, we propose an alternative measure: \emph{the Bayes' capacity}. This measure has been used previously in the security and information-theory literature as an important robustness measure for information leakage. Specifically, it provides 
an upper bound on the information leaked by a system, and to the extent by which such information can be used in \emph{any} kind of attack \cite{Alvim20:Book}.



We verify the developed theory by experimentally comparing the Gaussian and VMF mechanisms in terms of model accuracy and ability to defend against reconstruction attacks. Additionally, we compare the Bayes' capacity measure for each mechanism against its reconstruction accuracy, confirming that it is a better metric than $(\epsilon,\delta)$-DP   for this kind of attacks. 

To the best of our knowledge, this is the first work to present a common analytical framework to compare DP and metric private mechanisms and to design corresponding experiments to showcase their performance in deep learning applications. Moreover, to the best of our knowledge, it is also the first work to apply Bayes' capacity measures to deep learning and to show that it is the natural measure for privacy risk of reconstruction attacks. \\

\noindent \textbf{Key contributions:}
\begin{enumerate}
    \item We determine how to convert the parameters of the metric private mechanism VMF into those of RDP. Given that the RDP parameters can be converted to the $\epsilon$ and $\delta$ of DP, our study shows that RDP can be used as a common metric to compare DP and metric private mechanisms. 
    
    \item Using the state-of-the-art results in the DP and RDP literature, we introduce methods for privacy budget accounting for the VMF mechanism 
    that captures the effects of subsampling on privacy amplification and the number of algorithm epochs on privacy composition in the context of ML. 
   
    \item We experimentally demonstrate that mechanisms that are equivalent under the RDP framework have in fact very different utility in terms of accuracy, as well as very different privacy protection against reconstruction attacks. In particular, RDP-equivalent Gaussian and VMF exhibit better accuracy and better protection, respectively. 
    

    As an alternative measure of defence against reconstruction attacks,  we propose the Bayes' capacity. Indeed, we show that the Bayes' capacity determines the maximum amount of information that the attacker can use to determine the original gradients, that constitute a key component of such attacks. We validate theoretically and empirically that the Bayes' capacity is a better measure of protection against reconstruction attacks than the parameters of DP and RDP.
\end{enumerate}
}
  

\section{Related Work}
\label{sec:lit-rev}



\emph{Converting between frameworks via R\'enyi differential privacy}
For a theoretical comparison of the VMF and Gaussian mechanisms in terms of privacy versus utility performance, we need a fair method for the conversion of their mechanism parameters into common privacy or utility metrics. 
A promising and versatile framework for such conversion is  R\'enyi differential privacy~\citep{mironov:2017:CSF}. It is capable of handling important implementation aspects such as privacy amplification through sub-sampling~\citep{ mironov2019r, zhu2019poission,balle2018privacy,wang2019subsampled} and composition through multiple epochs ~\citep{abadi-etal:2016:CCS,Shahab_ISIT_2020:100rounds} of the SGD algorithm. Ultimately, the R\'enyi differential privacy computations will need to be translated back to the traditional $(\epsilon,\delta)$ differential privacy domain, where the first approach was provided in~\citep{mironov:2017:CSF}. Later works such as~\citep{Canonne_Kamath_Steinke_2022,Shahab_ISIT_2020:100rounds} have improved the conversion in the sense that tighter achievable $(\epsilon,\delta)$ tuples can potentially be obtained from the same R\'enyi differential privacy metric. For the Gaussian mechanism, these methods have been heavily studied~\citep{abadi-etal:2016:CCS, mironov2019r,Shahab_ISIT_2020:100rounds}. In this work, we use similar techniques to present how the R\'enyi differential privacy for the VMF mechanism is useful in characterising its privacy performance and how it enables a fair comparison of the Gaussian and VMF mechanisms. 

Another work that compares Gaussian and VMF mechanisms is \cite{faustini2023directional} in which the authors investigate this, like us, in the context of DP-SGD. Our work differs from theirs in that we focus on proposing a foundational framework to compare the different notions of formal privacy guarantees that these mechanisms provide rather than simply comparing them empirically via experiments. 
As far as we know, this is the only work that has performed this comparison; in general, there is a lack of studies comparing how different noise distributions affect the privacy/utility trade-off and how noise distributions other than isotropic ones can be used for training neural networks. 

\emph{Gradient-based reconstruction attacks.} For our experimental work, where we evaluate the effect on privacy, like~\cite{faustini2023directional}, we use gradient-based reconstruction attacks to calibrate how well each of our two mechanisms can provide protection against leaking of the original data, rather than Membership Inference Attacks (MIAs). This is because $a)$ reconstruction attacks are generally agreed in the literature to be more consequential breaches of privacy~\citep{melis-etal:2019:IEEE-SP, balle-etal:2022:IEEE-SP} and $b)$ \citep{faustini2023directional} showed that, although MIAs have been used for calibrating among standard DP variants \citep{DBLP:conf/uss/Jayaraman019}, they are not for calibrating between Gaussian and VMF mechanisms, while reconstruction attacks are.

The reconstruction attack we work with is a gradient-based one, where the attacker has access to these gradients.  
    Reference~\citet{zhu-etal:2019:NeurIPS} first developed the Deep Leakage from Gradients (DLG) attack, which can recover private data from neural network architectures which are twice differentiable. The attacker creates fake inputs, and by minimising the Euclidean distance between their gradients and the gradients received from another client, the fake input moves towards the real one.  An improved version was presented by \citet{DBLP:journals/corr/abs-2001-02610}.
    
    A subsequent method that improved over these was the Inverting Gradients method (\IGA) of \citet{geiping-etal:2020:NeurIPS}, which maximises the cosine similarity between gradients. It was proved in~\citet{geiping-etal:2020:NeurIPS}  that this reconstruction attack is guaranteed to be effective against Multi-Layer Perceptrons (MLPs). 
    
There has been much work since then on developing new gradient-based attacks and on understanding the conditions under which they apply or can be mitigated, e.g., see \citep{huang-etal:2021:NeurIPS,wu-etal:2023:UAI:learning-invert,wang-etal:2023:AISTATS:reconstructing-provably}, with a useful survey provided by \citet{du2024sok}.
For this paper, we use \IGA of \citet{geiping-etal:2020:NeurIPS} as a well-established method that is known to apply to the neural network architecture we use for our experiments.

\emph{Alternative measures for privacy.} The information theory and security communities have long used leakage measures to understand the security properties of systems modelled as information-theoretic channels. Recent work in these communities has converged on common measures based on the $g$-leakage framework of Smith~\cite{Smith09,issa2019operational}. The Bayes' capacity measure emerges in both communities as a robust leakage measure for both security and privacy~\cite{Alvim20:Book,sibson1969information,issa2019operational}.~\footnote{The logarithm of the Bayes' capacity is also known as the Sibson mutual information of order $\alpha = \infty$.} 
 
 Reference~\citet{fernandes2022explaining} compared the Bayes' capacity measure to the $\epsilon$ of local differential privacy, showing that both provide robust upper bounds on leakage but describe different adversarial scenarios. Our work validates these differences experimentally.

Reference~\citet{cherubinclosed} used a Bayes' security measure --based on the existing measure of \emph{advantage}-- to measure the leakage of DP-SGD. They show that the Bayes' security is bounded by the total variation distance between the distributions of the gradients. They use this measure to evaluate the Gaussian mechanism's efficacy at protecting against membership inference attacks and attribute inference attacks. Their work differs from ours in that our measures are different (we use the Bayes' capacity measure), and our applications are different (we use our measure to compare different mechanisms against the same reconstruction attack).

Reference~\citet{guo2022bounding} proposed Fisher Information Leakage as a better measure for privacy leakage against reconstruction attacks than $\epsilon$-DP, which they note has been identified as relevant to membership inference attacks. Similar to our work they focus on semantic guarantees relevant to reconstruction. However, our work differs significantly from theirs in that our focus is on the comparison of disparate mechanisms.

The closest work to ours is that of~\citet{hayes2023bounding}, who also propose an alternative measure for protection against reconstruction threats using a probability measure that measures the success of a reconstruction adversary. Their conclusion aligns with ours -- that $(\epsilon, \delta)$ is not a strong predictor of reconstruction accuracy. Our work differs from theirs in that our focus is on comparing different mechanisms, whereas they compare the same mechanism across different algorithms.

\section{The Privacy Model}
\label{sec:model}

One of the most successful recently popularised standards for asserting formal privacy guarantees has been \emph{differential privacy (DP)}~\cite{dwork-roth:2014,Dwork-etal:2006:DP} which ensures that a query output for a dataset remains probabilistically indistinguishable whether a specific personal record is contained in it or not. 
\begin{definition}[Differential privacy~\cite{dwork-roth:2014,Dwork-etal:2006:DP}]\label{def:DP}
For $\epsilon\in\mathbb{R}_{\geq 0}$ and $\delta\in[0,1)$, a mechanism $\mathcal{M}$ is $(\epsilon,\delta)$-differentially private if for all adjacent
datasets\footnote{Datasets are said to be \emph{adjacent} when they differ in one record.} $x$ and $x'$, denoted by $x\sim x'$,
and for every measurable $S\subseteq\operatorname{Range}(\mathcal{M})$, we have
\begin{equation} \label{eq:dp}
  \operatorname{Pr}\left[\mathcal{M}(x)\,\in\,S\,\right] \leq
  e^{\epsilon}\operatorname{Pr}\left[\mathcal{M}(x')\,\in\,S\right] + \delta
\end{equation}
\end{definition}

\begin{definition}[Metric privacy~\citep{chatzikokolakis-etal:2013:PETS}]\label{def:metric}
For an arbitrary set of secrets $\mathcal{X}$, equipped with a metric $d_{\mathcal{X}}$,  a mechanism $\mathcal{M}$ satisfies metric privacy or metric privacy if for all secrets $x$ and $x'$ for every measurable $S\subseteq\operatorname{Range}(\mathcal{M})$, we have
\begin{equation} \label{eq:metric}
  \operatorname{Pr}\left[\mathcal{M}(x)\,\in\,S\,\right] \leq
  e^{d_{\mathcal{X}}(x,x')}\operatorname{Pr}\left[\mathcal{M}(x')\,\in\,S\right]
\end{equation}  
\end{definition}

We start by briefly reviewing the probability density functions associated with the two mechanisms of interest, namely the Gaussian mechanism~\cite{Dwork-etal:2006:DP} that is regarded as the state-of-the-art to achieve DP in the context of machine learning (cf. Algorithm~\ref{alg:sgd}), and the von Mises-Fisher (VMF) mechanism~\cite{weggenmann-kerschbaum:2021:CCS} that has recently been in the spotlight for the formal guarantees it entails capturing the essence of the \emph{directional distance} between secrets (cf. Algorithm~\ref{alg:sgd2}). 

\subsubsection{Notation}

In the subsequent sections, we shall assume that our secrets are $p$-dimensional vectors and denote the space of all secrets as $\mathcal{X}\subseteq \mathbb{R}^p$. We write $\mathbb{S}^{p-1}$ to denote the unit sphere in $\mathbb{R}^p$ and $\mathbb{B}_R^{p}$ to denote the ball of radius $R$ in $\mathbb{R}^p$. That is, $\mathbb{S}^{p-1}=\{x\in\mathbb{R}^p\colon \|x\|_2=1\}$ and $\mathbb{B}_R^{p}=\{x\in\mathbb{R}^{p}\colon\|x\|_2 \leq R\}$. For a randomised function $f$ that associates to each element of $\mathcal{X}$ a distribution over $\mathcal{Y}$, we write $f(x)(y)$ for the value of the probability density function $f(x)$ at $y$.



\subsubsection{Gaussian mechanism}
The Gaussian mechanism is one of the most widely used DP mechanisms in the context of ML and is usually defined as follows.

\begin{definition}[Gaussian mechanism]\label{def:gaussian}
For any secret $x\in\mathcal{X}$ and and output $y\in\mathbb{R}^p$, the probability density function associated with the $p$-dimensional Gaussian mechanism with variance $\sigma^2$, denoted by $f_{p,\sigma^2}^G$, is given by
\begin{equation}\label{eq:gaussian}
f_{p,\sigma^2,x}^G(x)(y) ~ = ~ \frac{\exp{\left(-\frac{\|y-x\|_2^2}{2\sigma^2}\right)}}{\sqrt{2\pi\sigma^2}^p}
\end{equation}
\end{definition}

In the context of machine learning, the query associated with the Gaussian mechanism is typically the local SGD updates on the training data in each round and, in order to achieve a bounded sensitivity, the gradient updates are typically clipped in each round. Reference~\cite{abadi-etal:2016:CCS} was one of the pioneering works that formalised and delineated these notions with the incorporation of DP into ML.

\subsubsection{VMF mechanism}
Reference~\citet{weggenmann-kerschbaum:2021:CCS} introduced the VMF mechanism derived from the VMF distribution that perturbs an input vector $x$ on $\mathbb{S}^{p-1}$. They show that this mechanism satisfies $\epsilon d_\theta$-privacy, where $d_\theta$ is the angular distance between vectors, and $\epsilon$ is the scaling parameter corresponding to the $\kappa$ parameter in the definition of the VMF. 

\begin{definition}[VMF mechanism]
For any secret $x \in \mathbb{S}^{p-1}$ and an output $y\in\mathbb{S}^{p-1}$, the corresponding probability density function of the VMF mechanism centred around $x$ with a concentration parameter $\kappa > 0$ is given by

\begin{equation}\label{eq:VMF}
    f_{p,\kappa}^V(x)(y)~ = ~ \frac{1}{C_{p,\kappa}}e^{\kappa x^T y}~,
\end{equation}
where superscript $T$ denotes vector transpose, $C_{p,\kappa} = \frac{(2\pi)^{\nu+1}I_\nu(\kappa)}{\kappa^\nu}$ is the normalisation factor, $\nu = p/2-1$, and $I_\nu$ is the modified Bessel function of the first kind of order $\nu$. 
\end{definition}

The concentration parameter $\kappa$ roughly corresponds to the variance of the VMF around its mean; a larger $\kappa$ corresponds to a smaller variance.

\subsection{$(\epsilon, \delta)$-DP guarantees for the VMF mechanism}\label{sec:eps:guarantees}

As noted earlier, the VMF and Gaussian mechanisms provide seemingly incomparable guarantees, and thus their $\epsilon$ values cannot be compared directly. 
In order to compare the $(\epsilon, \delta)$-DP performance of Gaussian and VMF mechanisms, we employ the R\'enyi differential privacy (RDP) framework and the moments accounting (MA) methodology. That is, we first compute the respective RDP of Gaussian and VMF mechanisms and use the MA methodology to work out what $(\epsilon, \delta)$-DP they can achieve in DP-SGD after certain algorithm epochs, while also taking the effects of sub-sampling and privacy amplification into account. The Gaussian mechanism has been widely studied in the context of RDP and the reader is referred to~\citep{abadi-etal:2016:CCS, mironov2019r,Shahab_ISIT_2020:100rounds} for more details. Below, we introduce RDP for the VMF mechanism, which to the best of our knowledge has not been studied in the literature. 

\begin{definition}[R\'enyi differential privacy~\cite{mironov:2017:CSF}]\label{def:rdp}
A randomised mechanism $\mathcal{M}$ 
 satisfies $(\alpha,\tau)$-\emph{R\'enyi differential privacy (RDP)} if for any two adjacent datasets $x\sim x'$ we have
\begin{equation}\label{eq:rdp}
D_\alpha(\mathcal{M}(x)\|\mathcal{M}(x'))  \leq \tau,
\end{equation}
where $D_\alpha(P\|Q)$ denotes the R\'enyi divergence of order $\alpha$ of distribution $P$ from $Q$ and is given by
\[D_\alpha(P\|Q) = \frac{1}{\alpha -1} \log\left[\mathbb{E}_Q\left[\left(\frac{P}{Q}\right)^\alpha\right]\right].\]
\end{definition}

The R\'enyi divergence for two 
VMF distributions on $\mathbb{S}^{p-1}$ was derived in~\citep{kitagawa2022mises}. We use the specialised version of Proposition 3.1 in~\citep{kitagawa2022mises} for the case where the concentration parameters of both VMF distributions are equal to $\kappa$ and their centres are at antipodal 
directions resulting in the largest RDP quantity\footnote{
It can be proven that for any order $\alpha$, the R\'enyi divergence of two VMF distributions with the same concentration $\kappa$ is an increasing function of the angle between their centres $x$ and $x'$. Hence, the maximum divergence occurs when $x = -x'$.} to derive the following result. 
\begin{proposition}\label{prop:vmf:rdp}
Let $p \geq 2$ and $\nu = p/2-1$. For $\kappa \geq 0$ and $\alpha \in (1,\infty)$, the VMF mechanism satisfies $(\alpha, \tau_\alpha)$-RDP, where 
\begin{align}
\tau_\alpha = \frac{\nu}{\alpha-1}\log\left(\frac{1}{2\alpha-1}\right)+\frac{1}{\alpha-1}\log\left(\frac{I_\nu((2\alpha-1)\kappa)}{I_\nu(\kappa)}\right).\nonumber
\end{align}
The subscript $\alpha$ emphasises the dependence of RDP value on divergence order $\alpha$. For $\lim_{\alpha \rightarrow 1}$, the R\'{e}nyi divergence is computed as the KL divergence, which for the VMF mechanism satisfies $\tau_1 \leq 2\kappa \frac{I_{\nu+1}(\kappa)}{I_{\nu}(\kappa)}$. 
\end{proposition}

We have investigated the behaviour of the above RDP with respect to the VMF mechanism parameters $\nu$ and $\kappa$. We summarise our results in the following theorem.
\begin{restatable}{theorem}{VMFRenyiDivProp}\label{th:VMF_RenyiDiv_properties}
The Re\'nyi divergence $\tau_\alpha$ of the VMF mechanism for a given $\alpha\in(1,\infty)$ is:
\begin{enumerate}
    \item[$(i)$]an increasing function of $\kappa$ when $\kappa\geq 1$.
    \item[$(ii)$]a decreasing function of $\nu$.
\end{enumerate}
The proof is postponed to Appendix~\ref{app_proof: VMF_RenyiDiv_properties}.
\end{restatable}


For multi-layer neural networks, let us also consider a multi-variate VMF mechanism. Denote $\underline{p} = [p_1, \cdots, p_R]$, $\underline{\kappa} = [\kappa_1, \cdots, \kappa_R]$, $\underline{x} = [x_1, \cdots, x_R]$, and $\underline{y} = [y_1, \cdots, y_R]$, where $R$ is the number of VMF variables. The pdf of an independent multi-variate VMF mechanism is given by
\begin{equation}\label{eq:VMFmulti}    f_{\underline{p},\underline{\kappa}}^V(\underline{x})(\underline{y})~ = ~ \prod_{i=1}^{R}\frac{1}{C_{p_i,\kappa_i}}e^{\kappa_i x_i^T y_i}~.
\end{equation}
We can extend Proposition~\ref{prop:vmf:rdp} as follows.
\begin{proposition}\label{prop:vmf:rdp:multi}
For $i = 1, \cdots, R$, let $p_i \geq 2$ and $\kappa_i \geq 0$ be given and let $\nu_i = p_i/2-1$. For $\alpha \in (1,\infty)$, the independent multi-variate  VMF mechanism satisfies $(\alpha, \tau_\alpha)$-RDP where 
\begin{align}\label{eq:tau:alpha:multi}
\tau_\alpha = \frac{\sum\limits_{i=1}^{R}\nu_i\log\left(\frac{1}{2\alpha-1}\right)+\log\left(\frac{I_{\nu_i}((2\alpha-1)\kappa_i)}{I_{\nu_i}(\kappa_i)}\right)}{\alpha-1}.
\end{align}
 The proof is postponed to Appendix~\ref{app_proof:vmf:rdp:multi}.
\end{proposition}

An important question in applying the VMF mechanism to DP-SGD is how to divide and feed the gradient vectors across different layers of the network into the mechanism. Specifically, let the total parameter size of the gradient of the network be $p$. The question is how to break a $p$-dimensional gradient vector $x$ into an $R$-variate $\underline{x} = [x_1, \cdots, x_R]$, where each $x_i$ has dimension $p_i$ (i.e., $\sum_{i=1}^R p_i = p$). We address this question for the special case where $\kappa_i = \kappa\,\forall\,i=1,\cdots, R$. 

\begin{proposition}\label{prop:flattenisbest}
Let $\kappa_i = \kappa$, $i=1,\cdots, R$. Then the smallest Re\'nyi divergence $\tau_\alpha$ of the multi-variate VMF mechanism is achieved for $R=1$ and $ p_1 = p$. The proof is postponed to Appendix~\ref{app_proof:flattenisbest}.
\end{proposition}

Therefore, the VMF mechanism will give the best RDP guarantee when the VMF noise is applied to the overall gradient vector treated as a single vector. That is, we need to first \emph{flatten} the gradient vector across all layers of the network into a single $p$-dimensional vector and then feed it into the VMF mechanism.


There are at least two approaches for converting the $(\alpha, \tau_\alpha)$-RDP guarantees to $(\epsilon, \delta)$-DP guarantees, while also taking into account privacy amplification via sub-sampling (batch processing) of the input data and also the number of rounds or epochs in the DP-SGD algorithm. In general, there is no universal winner among these two approaches and each may work better for a different set of system parameters (see Table~\ref{tab:kappaepsilon}). Below, we briefly explain each approach.
\subsubsection{First Approach}\label{sec:first}
At a high level, this approach first converts the $(\alpha, \tau_\alpha)$-RDP guarantee to $(\epsilon, \delta)$-DP guarantee and then takes into account privacy amplification and composition of rounds in the $(\epsilon, \delta)$ domain. There are many methods in the literature for the conversion of the $(\alpha, \tau_\alpha)$-RDP to $(\epsilon, \delta)$-DP. We choose to use the method proposed in~\citep[Proposition 2.11]{Canonne_Kamath_Steinke_2022}, which is repeated here for ease of reference.
\begin{proposition}[\cite{Canonne_Kamath_Steinke_2022}]\label{prop:rdp:delta}
Suppose a mechanism $\mathcal{M}$ satisfies $(\alpha,\tau_\alpha)$-RDP. Then $\mathcal{M}$ also satisfies $(\epsilon, \delta)$-DP where
\begin{align}\label{eq:delta:alpha}
\delta_\alpha = \frac{e^{(\alpha-1)(\tau_\alpha-\epsilon)}}{\alpha-1}\Big(1-\frac{1}{\alpha}\Big)^{\alpha}.
\end{align}
By finding the infimum over $\alpha \in (1,\infty)$ we get the smallest $\delta$ (i.e., $\delta = \inf_{\alpha \in (1,\infty)}\delta_\alpha$).
\end{proposition}
This method gives close to optimal bounds in a numerically stable manner as compared with other works in the literature (e.g.,~\citep{Shahab_ISIT_2020:100rounds}). 

In fact, using standard techniques it can be proved that for the VMF mechanism with  $\tau_\alpha$ as given in Proposition~\ref{prop:vmf:rdp}, $\delta_\alpha$ is a convex function of $\alpha$. Taking the derivative of $\delta_\alpha$ with respect to $\alpha$ and setting it to zero, the optimal $\alpha^*$ is the one satisfying
\[-\epsilon + 2\kappa\frac{I_{\nu+1}((2\alpha^*-1)\kappa)}{I_{\nu}((2\alpha^*-1)\kappa)}+\log(1-\frac{1}{\alpha^*})= 0,\]
with the corresponding minimum $\delta$
\[\delta =  \frac{e^{(\alpha^*-1)(\tau_{\alpha^*}-\epsilon)}}{\alpha^*-1}\Big(1-\frac{1}{\alpha^*}\Big)^{\alpha^*},\]
where for any given $\epsilon$, $\alpha^*$ can be found numerically, e.g., using the bisection method.

Once $(\epsilon, \delta)$ is found, we take into account the effect of (Poisson) sub-sampling in privacy amplification~\citep[Theorem 8]{balle2018privacy} as follows:
\begin{align}\epsilon_s &= \log(1+\gamma(e^\epsilon-1)), \quad \delta_s = \gamma \delta,
\end{align}
where subscript $s$ stands for sub-sampling and $\gamma$ is the sampling probability, which is given by the ratio of the batch size to the dataset size. Using $(\epsilon_s, \delta_s)$, the last step is to take into account the privacy composition of multiple epochs of the algorithm. For this, we use~\citep[Theorem 10]{composition:theorems} which for completeness, is stated here using our notation.
\begin{theorem}[\citep{composition:theorems}]
For any $\epsilon_s >0$, $0\leq \delta_s < 1$ and $0\leq\tilde\delta < 1$, the class of $(\epsilon_s, \delta_s)$-DP mechanisms satisfies $(\tilde\epsilon_{\tilde\delta}, 1-(1-\delta_s)^N(1-\tilde\delta))$-DP under $N$-fold composition where
\begin{align*}
\tilde\epsilon_{\tilde\delta} &= \min\left\{N\epsilon_s,\dfrac{e^{
\epsilon _{s}}
-1}
{e^{\epsilon _{s}}+1}+\epsilon _{s}\sqrt{2N\log \left( e+\dfrac{\sqrt{N\epsilon _{s}^{2}}}{\tilde{\delta}}\right) },\right.\\&\qquad\qquad\left.\dfrac{e^{
\epsilon _{s}}-1}
{e^{\epsilon _{s}}+1}+\epsilon _{s}\sqrt{2N\log\dfrac{1 }{\tilde{\delta}}}\right\}
\end{align*}
\end{theorem}
For a given sampling rate $\gamma$, the number of epochs $N$, and a target $\delta_t$, one can vary the non-sub-sampled $\epsilon$ in a suitable range to obtain a range of $\delta$ and their corresponding  $(\epsilon_s, \delta_s)$ tuples, to then find the best triple $(\epsilon_s^*, \delta_s^*,\tilde{\delta^*})$ that results in the lowest $\tilde{\epsilon^*}$. 

\subsubsection{Second Approach}\label{sec:second} At a high level, this approach keeps the computations in the RDP domain for as long as possible, by taking into account the effects of sub-sampling and composition in the RDP domain and then converting to $(\epsilon, \delta)$-DP in the very last step.

We use the method presented in~\citep[Theorem 5]{zhu2019poission} for computing the sub-sampled Poisson RDP, which is repeated here using our notation.


\begin{theorem}[\citep{zhu2019poission}]
Let mechanism $\mathcal{M}$ satisfy $(\alpha,\tau_\alpha)$-RDP and let $\gamma$ be the sub-sampling probability. Then for any $\alpha\in \mathbb{Z}_{\geq 2}$, the sub-sampled mechanism satisfies $(\alpha,\tau^s_\alpha)$-RDP such that:
\begin{align*}
&\tau^s_\alpha \leq\frac{1}{\alpha}\log\Big\{(1-\gamma)^{\alpha-1}(\alpha\gamma\!-\!\gamma+1)\nonumber\\&+{\alpha\choose 2} \gamma^2(1-\gamma)^{\alpha-2}e^{\tau_2}\\&+3\sum_{\ell = 3}^\alpha{\alpha\choose \ell}(1-\gamma)^{\alpha-\ell}\gamma^\ell e^{(\ell-1)\tau_\ell}
\Big\},
\end{align*}
where superscript $s$ is used to denote sub-sampling. 
\end{theorem}

The above theorem does not provide the result for the special case of $\alpha = 1$ (which corresponds to the KL divergence of the mechanism). Using the convexity of the KL divergence and following similar lines of proof in~\citep[Theorem 5]{zhu2019poission}, it can be proved that for $\alpha = 1$, $\tau_1^s \leq \gamma \tau_1$.

 For non-integer values of $\alpha$, we can interpolate the RDP values~\citep[Corollary 10]{wang2019subsampled}. That is, for any $\alpha \in (1, \infty)$ we have
\[\tau^s_\alpha \leq (1-\alpha+\lfloor\alpha\rfloor)\tau^s_{\lfloor\alpha\rfloor}+(\alpha-\lfloor\alpha\rfloor)\tau^s_{\lceil\alpha\rceil}.\]

The effect of running $N$ epochs of the SGD algorithm on the sub-sampled RDP values is simply captured via multiplication by $N$~\citep{mironov:2017:CSF}. That is, $\tau'_\alpha = N\tau^s_\alpha$.


Finally, after obtaining $\tau'_\alpha$ as described above, we use Proposition~\ref{prop:rdp:delta} to convert $(\alpha, \tau'_{\alpha})$-RDP into $(\epsilon, \delta)$-DP. Often in practice, a target $\delta$ is desired. Hence, we can modify Eqn~\eqref{eq:delta:alpha} to instead express $\epsilon$ as a function of $\alpha$, $\delta$, and $\tau_\alpha$ as follows:
\[\epsilon = \frac{\alpha}{\alpha-1}\log\left(1-\frac{1}{\alpha}\right)+\tau_
\alpha - \frac{1}{\alpha-1}\log\left((\alpha-1)\delta\right)\]
This can be computed across a range of $\alpha$ to find the smallest possible $\epsilon$ achievable.

Table~\ref{tab:kappaepsilon} gives examples of  achievable $(\epsilon,\delta)$-DP computed for a range of $\kappa$. We used $\delta = 1/60000 \approx 1.67\times 10^{-5}$, which is the inverse of the dataset size used in our experiments, and calculated the best $\epsilon$ achievable through approach 1 or 2.  We also set $p = 13,700$, which is the number of parameters (size of the gradient vector) in the learning model in the experiments. We also use the sub-sampling probability of $\gamma = 128/60000$ and number of epochs $N=3$ to be consistent with the experiments. For more numerical examples and corresponding $\sigma$ in the Gaussian mechanism, refer to Appendix~\ref{app:full_results}.


\begin{table}
    \centering
    \begin{tabular}{|c|c|c|}
        \hline
        \textbf{$\kappa$ in VMF} & \textbf{$\epsilon$} & \textbf{Approach} \\
        \hline
        25 & 0.0139 &  1 \\
        \hline
        50 & 0.0867  & 1 \\
        \hline
        75 & 0.49 & 1 \\
        \hline
        100 & 2.5  & 2 \\
        \hline
        125 & 4.6 & 2 \\
        \hline
        150 & 7.97  & 2 \\
        \hline
        200 & 10.9  & 2 \\
        \hline              300 & 41.02  & 2 \\
        \hline
    \end{tabular}
    \caption{An example of whether the \emph{first} or the \emph{second} approach gives a better achievable $\epsilon$ for the VMF mechanism.}
    \label{tab:kappaepsilon}
\end{table}

\section{Privacy Against Reconstruction Attacks}
\label{sec:reconstruction}


While $(\epsilon, \delta)$-DP guarantees are a well-accepted measure for \emph{membership inference} attacks, in this paper we are also concerned with \emph{reconstruction attacks} which occur in the context of federated learning. In this section, we introduce a formal model for reconstruction attacks and demonstrate that Bayes' capacity -- a well-known measure from the information-theory and quantitative information flow literature -- aligns with the threat posed by a reconstruction attack. A primary version of such a model was introduced in \cite{biswas2024bayes}. We propose Bayes' capacity as an alternative robust measure for comparing mechanisms wrt.\ reconstruction threats. 


\subsection{Overview of the  attack}\label{sec:outline}


The reconstruction attacks we consider are \emph{gradient-based} reconstruction attacks, which describe an attacker with knowledge of the machine learning architecture, and who can observe the weight updates performed in the stochastic gradient descent algorithm. Significant works in this regard are  \emph{deep leakage from gradients} by Zhu et al.~\cite{zhu-etal:2019:NeurIPS}, \emph{see through gradients} by Yin et al.~\cite{yin-etal:2021:CVPR} and \emph{inverting gradients} by Geiping et al.~\cite{geiping-etal:2020:NeurIPS}. Such attacks are feasible in distributed learning architectures such as Federated Learning (FL), in which a single machine learning model is trained by a set of distributed machine learners, who send their gradient updates to a central server which performs weight aggregation and re-shares the global model. The threat model used in these works describes an honest-but-curious adversarial central server which can observe the weight updates of each of the machine learners in the system, and using knowledge of the machine learning architecture and the randomness in the stochastic gradient descent algorithm, can exactly reconstruct the input data.

An important caveat regarding the reconstruction attacks described in the literature is that they assume that no privacy protections have been put in place; that is, they do not consider an implementation of stochastic gradient descent with differential privacy, as in \Alg{alg:sgd}. 

The DP-SGD algorithm is usually tuned by adjusting the $(\epsilon, \delta)$-DP guarantees and comparing the accuracy obtainable from machine learner on real (non-training) data. In this case the $(\epsilon, \delta)$-DP guarantees are a proxy for the privacy protections provided by the machine learner. This should, in principle, include reconstruction attack protections. It is generally accepted that DP-SGD \emph{does} provide some protections against these attacks, however this has not, to our knowledge, been proven, nor has the extent of these protections been formally analysed.

The standard DP-SGD algorithm is depicted in \Alg{alg:sgd}. In federated learning, DP-SGD is performed in a distributed fashion: each client begins with their own training examples (inputs $x_1,...,x_N$) and a common loss function ($\mathcal{L}$); the server performs a weight initialisation (line 2) which is shared with the clients, who each perform the gradient update steps over their training batch (lines 4-10) and then send back the gradients to the server for the gradient descent step (line 12). The updates ($\theta_{t+1}$) are shared with the clients for the next ($t+1$) round. The leak statement (line 11) indicates the information observed by the adversary, who in this scenario is the server. We remark that the standard SGD algorithm (without differential privacy) excludes lines 7 (clipping) and 10 (noise addition).

\begin{algorithm}[!t]
\caption{DP-SGD with Gaussian noise}\label{alg:sgd}
\begin{algorithmic}[1]
\State \textbf{Input:} Examples $\{x_1,\ldots,x_N\}$, loss function $\mathcal{L}(\theta) = \frac{1}{N} \sum_i \mathcal{L}(\theta, x_i)$. Parameters: learning rate $\eta_t$, noise scale $\sigma$, group size $L$, gradient norm bound $B$.
\State \textbf{Initialise} $\theta_0$ randomly
\For{$t \in T$}
   \State $L_t \gets $ random sample of $L$ indices from $1{\ldots}N$
   \For{$i \in L_t$}
   	\State $\mathbf{g}_t(x_i) \gets  \nabla_{\theta_t} \mathcal{L}(\theta_t, x_i)$
    \Comment{Compute gradient}
	\State $\gbar_t(x_i) \gets \nicefrac{\mathbf{g}_t(x_i)}{\max (1, \frac{\| \mathbf{g}_t(x_i)\|_2}{B}) }$
\Comment{Clip gradient}
  \EndFor  
   \State $\gtilde_t \gets \frac{1}{L} \sum_i \gbar_t(x_i)$
   \Comment{Average}
   \State $\gtilde_t \gets \gtilde_t + \frac{1}{L} \Gauss(0, B^2\sigma^2)$
   \Comment{Add noise}
   {\color{red}\State // Leak $\gtilde_t$}
   \State $\theta_{t+1} \gets \theta_t - \eta_t \gtilde_t$
    \Comment{Descent}
\EndFor
\State \textbf{Output} $\theta_T$ 
\end{algorithmic}
\end{algorithm}

The reconstruction goal of the adversary is to learn the exact value of the training examples of a client (chosen from $x_1,...,x_N$ in line 4), using their knowledge of the model, the initial parameters $\theta_0$, the loss function $\mathcal{L}$, the size of the batch $L$ and the leaked gradient $\gtilde_t$. 

\subsection{Model for federated learning with DP-SGD}\label{sec:formal_model}
Formally, we model a system as an information-theoretic channel $C:\calx \rightarrow \Dist{\caly}$, taking secrets $\calx$ to distributions over observations $\caly$ can be written as a matrix $C:{\calx \times \caly}\rightarrow [0,1]$ whose rows are labelled with secrets, and columns labelled with observations, and where $C_{x, y}$ is the probability of observing $y \in \caly$ given the secret $x \in \calx$. 

Applying this model to \Alg{alg:sgd}, we have that each client can be expressed as a channel taking a set of $L$ inputs $x_L \in \calx^L$ (line 4) and producing an observable (noisy) averaging-of-gradients $y \in \caly$ (line 10), where we denote by $\caly$ the set of all possible observations and $\calx^L$ the set of all possible input sets of length $L$. For now our model assumes that $\calx^L$ and $\caly$ are discrete; we extend this idea to continuous domains in \Sec{sec:bayes}.  We remark that once the set of size $L$ is selected, the algorithm consists of deterministic steps (lines 5-9), followed by a probabilistic post-processing (line 10). We can model this system as a composition of channels $D \circ C$, where $C$ is the deterministic channel described by lines 5-9 and $D$ is the noise-adding mechanism from line 10. We write the type of $D$ as 
$\caly \rightarrow \Dist{\calz}$, that is, taking vectors to distributions over noisy vectors, and the type of $C$ as $\calx^L \rightarrow \Dist{\caly}$. Note that $\Dist{\caly}$ incorporates deterministic channels as point distributions on $\caly$.  Mathematically, the composition $D \circ C$ corresponds to $C {\cdot} D$ where $\cdot$ is matrix multiplication. This composition describes the information flow to the attacker, from secrets $\calx^L$ to observations ${\calz}$. 

\subsection{Model for the reconstruction attack}

We model a Bayesian attacker: they are equipped with a prior over secret inputs, modelled as a probability distribution $\pi: \Dist{\calx}$. Usually, the choice of prior is significant, as it determines the overall leakage of the system to an adversary. Here we turn to the empirical literature on attacks. To increase the effectiveness of gradient inversion attacks, a \emph{strong prior} can be used as a regulariser to ensure that the model favours more realistic images~\cite{yin-etal:2021:CVPR}. The effect of this is that the space of potential secrets (images) is reduced to a set of ``feasible'' images which could have produced the observed gradient. 
That is, the effect is to limit the support of the prior on images. However, we assume that there is no reason for the adversary to favour any particular feasible image (or set of images) which remain in the support.  
Thus, in our theoretical analysis, we model a prior which is \emph{uniform over its support}.
We discuss the limitations of this approach in \Sec{sec:results-bayes}.


We equip our attacker with a gain function $g:{\calw \times \calx} \rightarrow \Reals$ which describes the gain to the attacker when taking action $w \in \calw$ if the real secret is $x \in \calx$. In our reconstruction attack model for FL, the adversary's goal is to learn the secret (training input) \emph{exactly}, using a single observation of the gradient updates, which corresponds to the following gain function describing an attacker who learns the secret exactly in one guess (called a one-try attack). 

\begin{equation}\label{eqn:gain}
	\Bayes(w, x) ~=~ \left\{
		\begin{array}{ll}
			1 & \text{ if $x = w$,} \\
			0 & \text{ otherwise.}
		\end{array}
	\right.
\end{equation}


We model the prior vulnerability of the system as the attacker's maximum expected gain using only their prior and gain function:
$
     \VgPrior{\Bayes}(\upsilon) = \max_{x \in \supp{\calx^L} } \upsilon_x = \frac{1}{|\supp{\calx^L}|}~,
$
where $\supp{\pi}$ denotes the support of the distribution $\pi$.

Our attacker is assumed to have knowledge of the machine learning architecture and the DP-SGD algorithm described in \Alg{alg:sgd}. As explained above, this system can be modelled as the composition $C{\cdot}D$. Thus, our attacker's posterior knowledge can be computed as their maximum expected gain after access to the channel: 
\begin{equation}
    \VgPosterior{\Bayes}(\upsilon, C{\cdot}D) =  \frac{1}{|\supp{\calx^L}|} \sum_{z \in \calz} \max_{x \in \supp{\calx^L}} (C{\cdot}D)_{x, z}.
\end{equation}
%
The leakage of the secrets via the channel $C{\cdot}D$ to the adversary is then the (multiplicative) difference between the prior and posterior vulnerabilities:
\begin{equation}\label{eqn:dpsgd-leak}
   \gLeak{\Bayes}(\upsilon, C{\cdot}D) =  \sum_{z \in \calz} \max_{x \in \supp{\calx^L}} (C{\cdot}D)_{x, z}.
\end{equation}

\section{Bayes' Capacity}\label{sec:bayes}

The quantity computed in \Eqn{eqn:dpsgd-leak} is a measure known as the \emph{Bayes' capacity}~\cite{Alvim20:Book}. For a channel $M: \calx \rightarrow \Dist{\caly}$ it is defined:

\begin{equation}\label{eqn:bayes_capacity}
	\CBayes(M) ~=~ \sum_{y \in \caly} \max_{x \in \calx} M_{x, y}~.
\end{equation}

This quantity was shown to measure the maximum leakage of adversaries wanting to guess arbitrary values of the secret $\calx$. More in general, the Bayes' capacity is a measure of the maximum multiplicative leakage of a system to \emph{any} adversary modelled using a prior and a gain function (this result is known as the \emph{Miracle Theorem}~\cite{m2012measuring}). In the case of reconstruction attacks, this means that even if our adversarial assumptions are incorrect (eg. the adversary's prior is not uniform, or their reconstruction attack goal is to find a value close to the secret rather than the exact value of the secret), then the Bayes' capacity represents a tight upper bound on the leakage of the system to this adversary. 

The Bayes' capacity has the following properties (The proofs are postponed to Appendix~\ref{app:bayes_proofs}).

\begin{lemma}\label{lem:BC_prop_0}
Let $C$, $D$ be channels such that $C{\cdot}D$ is defined, and $C$ is deterministic with no $0$ columns. Then $\CBayes(C{\cdot}D) = \CBayes(D)$.
\end{lemma}

\begin{lemma}\label{lem:BC_prop_1}

 Denote by $M_\mathbf{1}$ the channel that leaks nothing, and by $M_\mathbf{0}$ the channel that leaks everything. Then $\CBayes(M_\mathbf{1}) = 1$ and $\CBayes(M_\mathbf{0}) = N$ where $N$ is the size of the domain of $M_\mathbf{0}$.
\end{lemma}

\noindent Note that the capacity increases with the size of the domain, which gives the following corollary.

\begin{corollary}\label{corr:size of secret set}
Let $C:\calx \rightarrow \Dist{\caly}$ and $X, X' \subseteq \calx$. Denote by $C_X$ the channel $C$ restricted to domain $X$, and similarly for $X'$. Then $X \subseteq X'$ implies $ \CBayes(C_{X}) \leq \CBayes(C_{X'})$.
\end{corollary}

The following is a version of the data processing inequality, which says that post-processing does not increase leakage.
\begin{lemma}\label{lem:BC_prop_2}
Let $C:\calx \rightarrow \Dist{\caly}$ and $D: \caly \rightarrow \Dist{\calz}$. Then $\CBayes(C{\cdot}D) \leq \CBayes(C)$.
\end{lemma}


Returning now to DP-SGD, we show how to simplify \Eqn{eqn:dpsgd-leak} using Bayes capacity, thereby defining reconstruction risk in terms of the noise-adding mechanism $D$. 

\begin{theorem}[Reconstruction risk for FL]\label{lem:reconstruction-risk}
Given a DP-SGD algorithm described by a deterministic pre-processing of inputs $C$ followed by a gradient-noise-adding mechanism $D$ such that the range of $C$ is the domain of $D$, then the reconstruction risk caused by the leakage of $C{\cdot}D$ is given by $\CBayes(D)$.
\end{theorem}

\Cref{lem:reconstruction-risk} implies that the Bayes' capacity of the noise-adding component of the DP-SGD algorithm is sufficient to measure the leakage of the entire algorithm wrt.\ reconstruction threats. Thus, finally, we can use the Bayes' capacities to compare our privacy mechanisms for safety against reconstruction threats. 


\begin{definition}[Safety against reconstruction]\label{def:safety}
 Let $\CBayes$ model the reconstruction threat against mechanisms $M$ and $M'$ defined on the same domain.
 We say that $M$ is safer than $M'$ against a reconstruction threat iff $\CBayes(M) < \CBayes(M')$.
\end{definition}

\subsection{Continuous Bayes' Capacity}

In \Sec{sec:reconstruction} we derived a Bayes' capacity formulation for reconstruction attacks using a composition $C{\cdot}D$, where $D$ represented the noise-adding components of DP-SGD. However, we did not provide a method for computing $D$. In the following we show how to compute the channel $D$ and the resulting Bayes' capacity for two mechanisms of interest: the Gaussian mechanism, described in \Alg{alg:sgd} and parametrised by $\epsilon$ and $\delta$ values, and the VMF mechanism, described in \Alg{alg:sgd2}, which satisfies metric differential privacy and is parametrised by $\epsilon$ alone. 

For the both mechanisms, the output domain of interest is continuous, however, Bayes' capacity has previously only been defined for mechanisms acting on discrete domains. We now introduce a natural generalisation of Bayes' capacity for continuous domains.

\begin{definition}[Bayes' capacity for continuous domains]
Let $f: \mathcal{X} \rightarrow \mathcal{Y}$ be a randomised function taking inputs $x \in \mathcal{X}$ to distributions on outputs $\mathcal{Y}$. Then the Bayes' capacity of $f$ is defined as 
\begin{equation}\label{eqn:bcapacity}
    \CBayes(f) ~=~ \int_{\mathcal{Y}} \sup_x f(x)(y)~ dy
\end{equation}
where $f(x)(y)$ denotes the (continuous) probability density $f(x)$ evaluated at $y$. This is well-defined when $f$ is measurable (since the pointwise supremum of measurable functions is measurable).
\end{definition}


\subsection{Bayes' capacity for Gaussian noise}

We can now derive the Bayes' capacity for the DP-SGD algorithm using Gaussian noise, described by lines 4-10 of \Alg{alg:sgd}. From \Lem{lem:BC_prop_0},
since lines 4-9 are deterministic, we simply calculate the Bayes' capacity for the operation defined in line 10. We set clipping length $C=1$ to align with our experiments.\footnote{In order to remain consistent with the DP-SGD literature, we use $C$ for clipping length, which is a scalar. This should not be confused with the deterministic channel $C$.}

\begin{theorem}[Bayes' Capacity for Gaussian noise]\label{th:BC_Gaussian}
    Let $G_{p, \sigma}: \mathcal{X} \rightarrow \mathbb{R}^{p}$ be the mechanism described by \Alg{alg:sgd} (line 10) which takes as input a $p$-dimensional vector $x \in \mathcal{X}$ and outputs a perturbed  vector $y \in \mathbb{R}^{p}$ by applying Gaussian noise with parameter $\sigma$ to each element of the (clipped) input vector with clipping length $C=1$, and then averaging by $L$. Then the Bayes' capacity of $G_{p, \sigma}$ is given by:
\[
    \CBayes(G_{p, \sigma}) ~=~ \frac{2}{\Gamma\left(\frac{p}{2}\right) 2^{\frac{p}{2}} \sigma^p} Z + \frac{R^p}{\Gamma\left(\frac{p}{2} + 1\right) 2^{\frac{p}{2}} \sigma^{p}}
\]
where $R$ is the number of layers in the network (each clipped to length 1) and $Z = \sum_{i=0}^{p-1} \Gamma(\frac{p-i}{2}) (\sqrt{2} \sigma)^{p-i} { p-1 \choose i } R^i$. The proof is postponed to Appendix~\ref{app_proof:BC_Gaussian}.
\end{theorem}

\subsection{Bayes' capacity for VMF noise}

\Alg{alg:sgd2} is a modified version of DP-SGD designed for von Mises-Fisher (VMF) noise. The blue text highlights differences from the Gaussian DP-SGD, and in particular we note that the VMF mechanism requires scaling of the final vector to ensure that it resides on the unit sphere $\mathbb{S}^{p-1}$. (Here $p$ is computed as the number of weights in the network).

As for the Gaussian mechanism, we find that the DP-SGD algorithm for VMF decomposes into the same deterministic channel $C$ composed with a channel $D$ described by lines 10-11 of \Alg{alg:sgd2}. We now compute the Bayes' capacity for the channel $D$ using the VMF mechanism.

\begin{algorithm}[!th]
\caption{DP-SGD with von Mises-Fisher noise}\label{alg:sgd2}
\begin{algorithmic}[1]
\State \textbf{Input:} Examples $\{x_1,\ldots,x_N\}$, loss function $\mathcal{L}(\theta) = \frac{1}{N} \sum_i \mathcal{L}(\theta, x_i)$. Parameters: learning rate $\eta_t$, noise scale $\sigma$, group size $L$, gradient norm bound $B$.
 \State \textbf{Initialise} $\theta_0$ randomly
 \For{$t \in T$}
    \State $L_t \gets $ random sample of $L$ indices from $1{\ldots}N$
    \For{$i \in L_t$}
    	\State $\mathbf{g}_t(x_i) \gets  \nabla_{\theta_t} \mathcal{L}(\theta_t, x_i)$
     \Comment{Compute gradient}
 	\State $\gbar_t(x_i) \gets \nicefrac{\mathbf{g}_t(x_i)}{\max (1, \frac{\| \mathbf{g}_t(x_i)\|_2}{B}) }$
  \Comment{Clip gradient}
   \EndFor  
  \State $\gtilde_t \gets \frac{1}{L} \sum_i \gbar_t(x_i)$  
   \Comment{Average}
   {\color{blue}  \State $\gtilde_t \gets \nicefrac{\gtilde_t}{ | \gtilde_t |}$ }
    \Comment{Scale}
    {\color{blue} \State $\gtilde_t \gets \VM(\sigma, \gtilde_t)$}
    \Comment{Add noise}
   {\color{red}\State // Leak $\gtilde_t$}
    \State $\theta_{t+1} \gets \theta_t - \eta_t \gtilde_t$
    \Comment{Descent}
 \EndFor
 \State \textbf{Output} $\theta_T$ 
\end{algorithmic}
\end{algorithm}



\begin{theorem}[Bayes' Capacity for VMF]\label{th:BC_VMF}
Let $V_{p, \kappa}:\mathcal{X} \rightarrow \mathbb{S}^{p-1}$ be the mechanism described by \Alg{alg:sgd2} (lines 10-11) which takes as input a $p$-dimensional vector $x \in \mathcal{X}$ and outputs a perturbed vector $y \in \mathbb{S}^{p-1}$ by applying VMF noise with parameter $\kappa$ on the unit sphere $\mathbb{S}^{p-1}$. Then the Bayes' capacity of $V_{p, \kappa}$ is given by
\[
    \CBayes(V_{p, \kappa}) ~=~ 2 \Gamma^{-1}\left(\frac{p}{2}\right) \frac{\kappa^{\frac{p}{2} - 1}}{2^{\frac{p}{2}} I_{\left(\frac{p}{2} - 1\right)}(\kappa)} e^{\kappa}
\]
where $I_\nu$ is the modified Bessel function of the first kind of order $\nu$.
The proof is postponed to Appendix~\ref{app_proof:BC_VMF}
\end{theorem}

\subsection{Comparing DP-SGD mechanisms using Bayes' capacity}\label{sec:comparing-bayes}

The reconstruction attack model for DP-SGD with Gaussian is $M = C{\cdot}G_{p, \sigma}$, where we write $C$ for the deterministic channel represented by lines 5-9 of \Alg{alg:sgd} followed by the Gaussian at line 10. For \Alg{alg:sgd2} with VMF the model is $M' = C{\cdot}V_{p, \kappa}$, with the same $C$ as for the Gaussian but now the VMF noise mechanism $V_{p, \kappa}$ represented by lines 10-11. By \Def{def:safety}, we can compare these algorithms wrt.\ reconstruction attacks by comparing $\CBayes(C{\cdot}G_{p, \sigma})$ and $\CBayes(C{\cdot}V_{p, \kappa})$. We can simplify this to a comparison of $\CBayes(G_{p, \sigma})$ and $\CBayes(V_{p, \kappa})$.

In other words, we have that \Alg{alg:sgd} is safer against reconstruction attacks than \Alg{alg:sgd2} iff $\CBayes(G_{p, \sigma}) < \CBayes(V_{p, \kappa})$ and vice versa. We will use this comparison in our experiments in \Sec{sec:results-bayes}.

\section{Experimental Setup}
\label{sec:exper}

We train classification neural networks with the VMF and Gaussian mechanisms and evaluate them across utility and privacy tasks. For utility, we compare the accuracy of a model trained without DP guarantees against the models trained under each mechanism. For privacy, we first measure how VMF and Gaussian prevent gradient-based reconstruction attacks for the same $\epsilon$ values. We broadly follow the experimental setup of \cite{faustini2023directional}.  However, we diverge in two respects.  First, to avoid potentially confounding effects from multiple gradients, and noting the remarks following Propn~\ref{prop:flattenisbest},
we use sufficiently small models and data to allow us to work with single flattened gradients.  Second, due to the theory earlier in the paper, the two mechanisms can be executed under the
same privacy parameter.

\paragraph{Datasets}\label{sec:datasets}
We use the standard training (60,000 images) and test sets (10,000 images) from \textbf{MNIST} \cite{deng2012mnist} and  \textbf{Fashion-MNIST} \cite{Xiao2017FashionMNISTAN} datasets. Each image belongs to one of ten classes. 
We downsized the datasets from 28x28 images to 16x16 images: we resized the Fashion-MNIST images but cropped the MNIST ones at the centre since most of the background of its images is mono-colour black.  Doing this allows us to straightforwardly apply the flattened gradient framework of the theory in this paper 
while still giving reasonable utility results on these datasets.

\paragraph{Model}
\label{sec:exper-baseline}
Our model is a Multilayer Perceptron (\textbf{MLP}), which is a feedforward neural network. We use sigmoid as the activation function and no bias vectors. The model has an input layer ($256 \times 50 = 12,800$ parameters), a hidden layer ($50\times 15=750$ parameters) and an output layer ($15\times 10 = 150$ parameters), summing 13,700 parameters. Reference~\cite{geiping-etal:2020:NeurIPS} proves that their IGA attack can uniquely reconstruct the input from the gradient against fully connected networks.  This is a relatively small network: we chose this configuration for the reasons noted above for datasets, as it allows us to straightforwardly work with flattened gradients.

For both datasets, we trained the models for 3 epochs using a batch size of 128. We used the Adam optimiser with a learning rate of 0.005 and weight decay of 0.1. We use the Opacus \cite{DBLP:journals/corr/abs-2109-12298} implementation of the Gaussian mechanism and integrate the VMF\footnote{https://github.com/dlwhittenbury/von-Mises-Fisher-Sampling} sampling with it in our code.

We run the Inverting Gradients Attack\footnote{https://github.com/JonasGeiping/invertinggradients} for 10,000 iterations and measure the mean-squared error (MSE) and the Structural Similarity Index (SSIM) \cite{1284395} scores between original and reconstructed images. 
SSIM has been shown to correlate with human judgements of images whereas MSE captures just the quantitative difference between images; both are widely used in estimating privacy leakage in reconstruction attacks as observed by e.g. \cite{sun-etal:2023:NeurIPS}.
For consistency with the utility experiments, we also use a batch of 128 images and the same amount of noise in terms of $\kappa$ for VMF and \emph{noise multiplier} $\sigma$ for Gaussian.

We also considered an extra experiment with the DLG attack \cite{sun-etal:2023:NeurIPS}, which is the seminal study on gradient-based reconstructions. However, as its authors report, it assumes many restrictions. Amongst them, it only works for batch sizes up to 8, which is too limited in practice.
Also, preliminary experiments within our setup showed that it failed to reconstruct under even the lightest noise of either mechanism, rendering it not useful for mechanism comparisons.
We therefore instead adopted the IGA attack. Its authors claimed that the input to any fully connected layer can be reconstructed, which suits well our (fully connected) MLP architecture. Moreover, their experiments attacked coloured images in a batch of 100 images, similar to our batch size.   

\section{Experimental Results}
\label{sec:results}

We present first the utility results, which are carried out on classification tasks on the test sets of MNIST and FashionMNIST datasets. We then report the privacy experiments, which analyse the effect of the mechanisms against image reconstruction from gradients. We then look at how utility and privacy compare under the relationship established by RDP analysis, and in Section~\ref{sec:results-bayes} we look at these through the lens of Bayes capacity.

               
        

    
                
        

\begin{figure*}[ht!]
    \centering
    \small
    \begin{tabular}{cc}
               
        \subfloat[MSE ]{\includegraphics[width=0.4\textwidth]{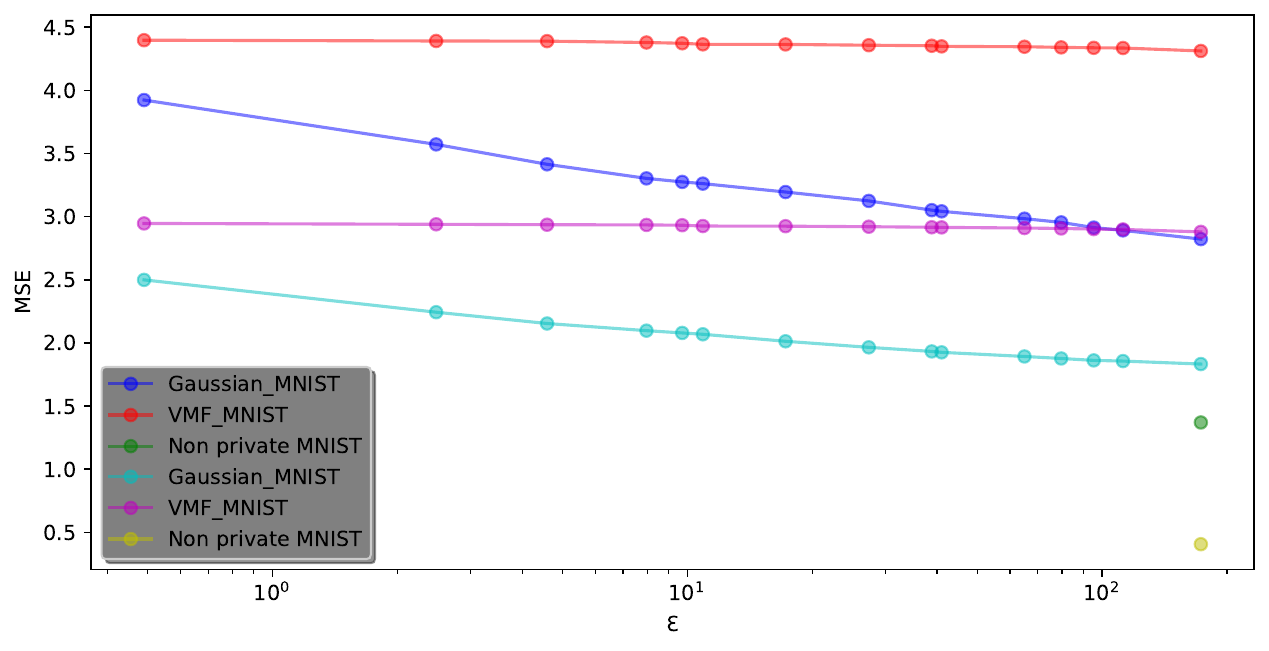}} &
        
        \subfloat[SSIM ]{\includegraphics[width=0.4\textwidth]{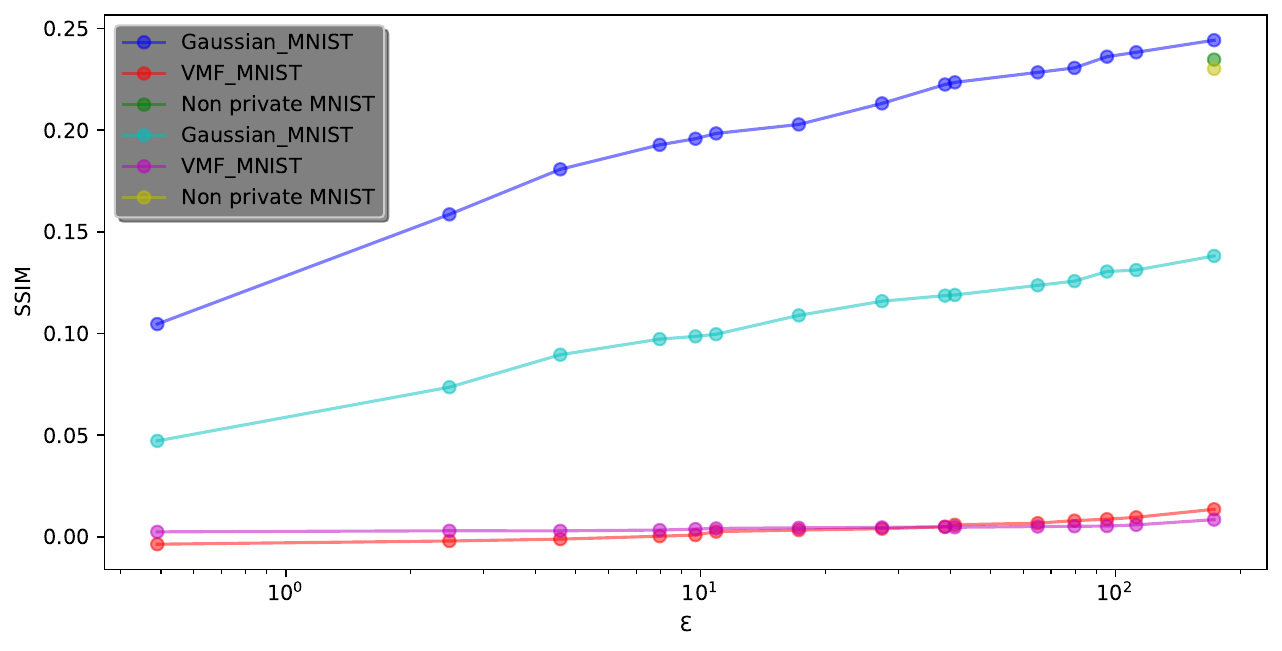}} \\       
        
    \end{tabular}
    \vspace{4mm}
    \caption{Reconstruction results, in terms of SSIM or MSE, across the datasets under different privacy settings by the Inverting Gradients Attack (IGA).}\label{fig:reconstruction_plots}
   \end{figure*}

\subsection{Utility}
\label{sec:results-utility}

\begin{figure*}[t!]
    \centering
    \small
    \begin{tabular}{cc}
               
        \subfloat[Accuracy for MNIST]{\includegraphics[width=0.4\textwidth]{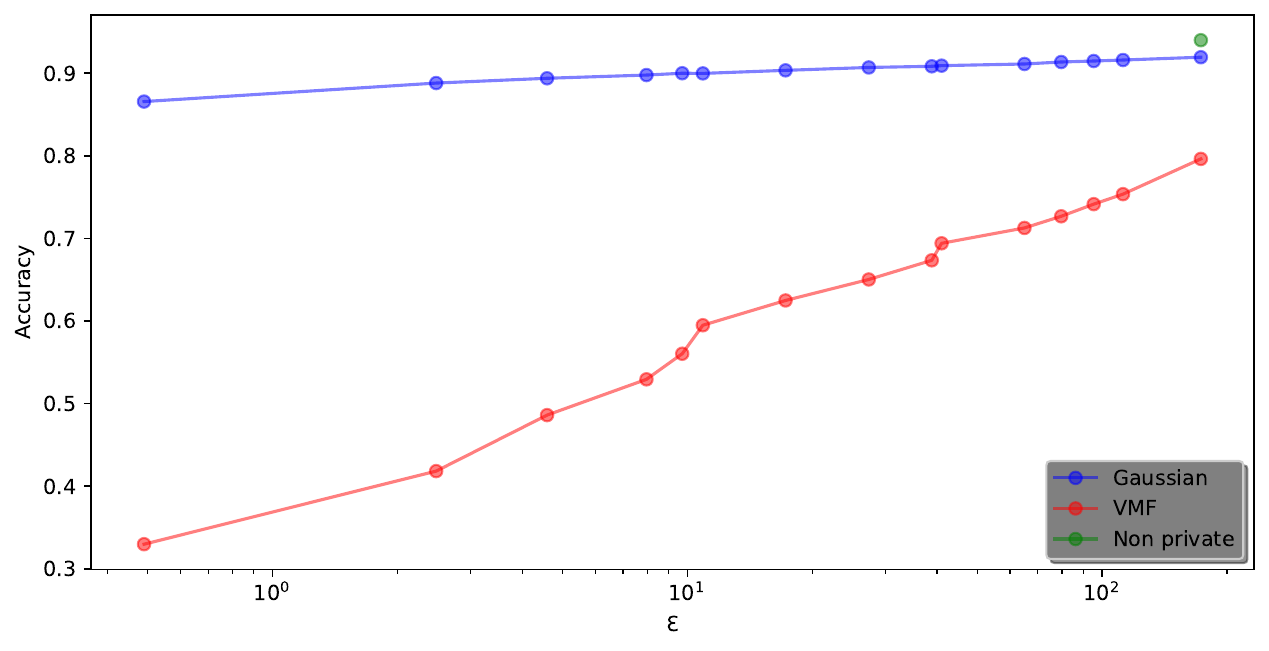}} &
        
        \subfloat[Accuracy for FashionMNIST]{\includegraphics[width=0.4\textwidth]{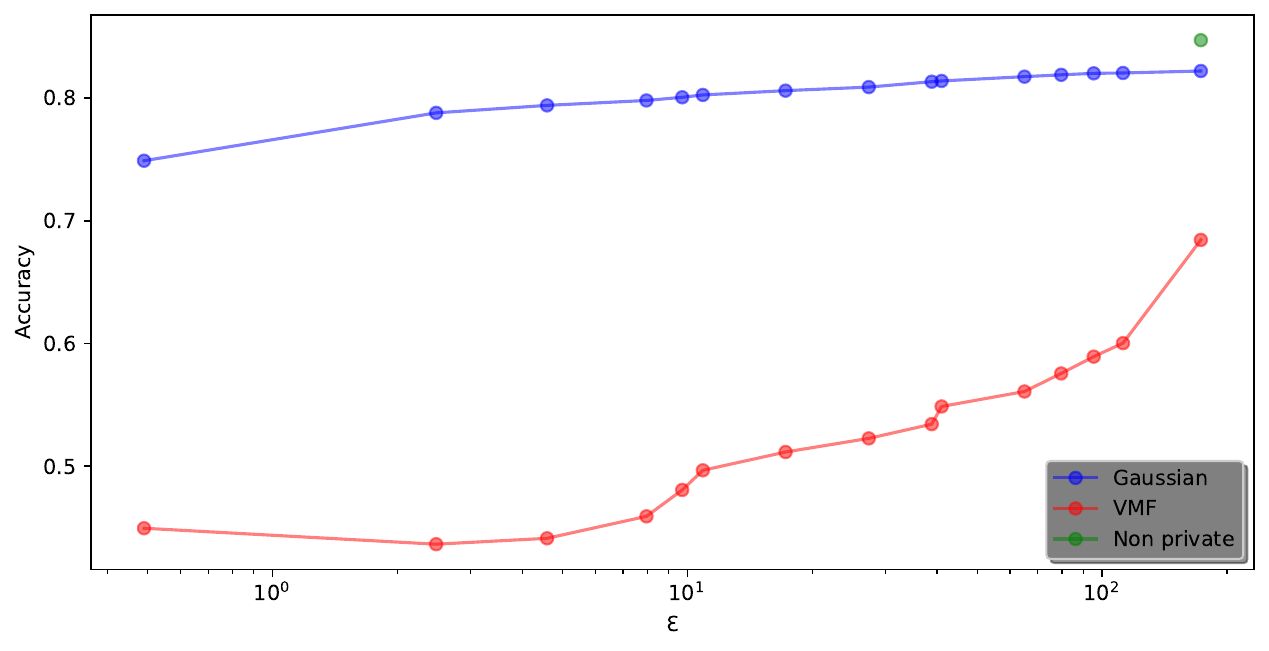}} 
    \end{tabular}
    \vspace{4mm}
    \caption{Utility results, in terms of accuracy, across the datasets under different privacy settings.}\label{fig:utility_plots}
\end{figure*}

We mapped the relationship we have established for $\kappa$ in Sec.~\ref{sec:first} and~\ref{sec:second} and for $\sigma$~\citep{abadi-etal:2016:CCS, mironov2019r,Shahab_ISIT_2020:100rounds} in terms of $\epsilon$ for up to $173$. In this range, there are no intersection points in terms of utility, as Figure~\ref{fig:utility_plots} shows; in this range, the Gaussian mechanism always outperforms VMF in both datasets. The non-private model, in turn, yields higher accuracy than any private model, despite the accuracy of the private model increasing with larger $\epsilon$. Still, models trained under either of the mechanisms, even for tight privacy budgets, achieve much higher accuracy than a random baseline which, for a 10-classes balanced problem, is expected to be 10\%. For instance, for $\epsilon = 0.49$, the accuracy for the Gaussian model is 86.56\% and 74.89\% for MNIST and FashionMNIST respectively, whereas for VMF it is 32.97\% and 44.95\%. The gap between the mechanisms drops as $\epsilon$ increases and the models approach the performance of the non-private baseline. 

We also note that cropping or resizing the images incur in a penalty in terms of accuracy that is achievable with state-of-the-art models, which in turn are also more complex than our 13,700-parameter neural network. Still, our non-private model achieves an accuracy of 93.99\% in MNIST (SOTA is 99.87\%\footnote{https://paperswithcode.com/sota/image-classification-on-mnist}) and 84.71\% in FashionMNIST (SOTA is 89.7\%\footnote{http://fashion-mnist.s3-website.eu-central-1.amazonaws.com/}).

\subsection{Privacy}~\label{sec:results-privacy}
    \begin{table}[h!]
        \centering
        \small
        \begin{tabular}{cc|cc|cc}
        \hline
        \textbf{Mechanism} &
        \textbf{\makecell{$\kappa$ (VMF)/ \\ $\sigma$ (Gauss)}    } &
        \multicolumn{2}{c}{\textbf{MNIST}} & \multicolumn{2}{c}{\textbf{FMNIST}} \\

        &&\textbf{SSIM}  & \textbf{MSE} & \textbf{SSIM}  & \textbf{MSE}  \\ \hline

        None&-&.2347 & 1.370&.2301 &.4070 \\ 
        None&Clip& .2577&2.637 & .1559&1.734 \\
        Gauss&1.23& .1046&3.923 &.0472&2.499 \\ 
        VMF&75& -.0035&4.396 & .0024&2.946 \\ 
        Gauss&.0174& .2443&2.822 & .1381&1.833 \\ 
        VMF&500& .0135&4.311 & .0085&2.879 \\ 
          
          \hline
         \end{tabular}
         \vspace{4mm}
        \caption{IGA reconstruction metrics for selected noises} 
        \label{tab:reconstruction_metrics}
    \end{table}

\begin{figure*}[ht!]
    \centering
    \begin{tabular}{cc}
               
        \subfloat[MNIST reconstructions]{\includegraphics[width=0.4\textwidth]{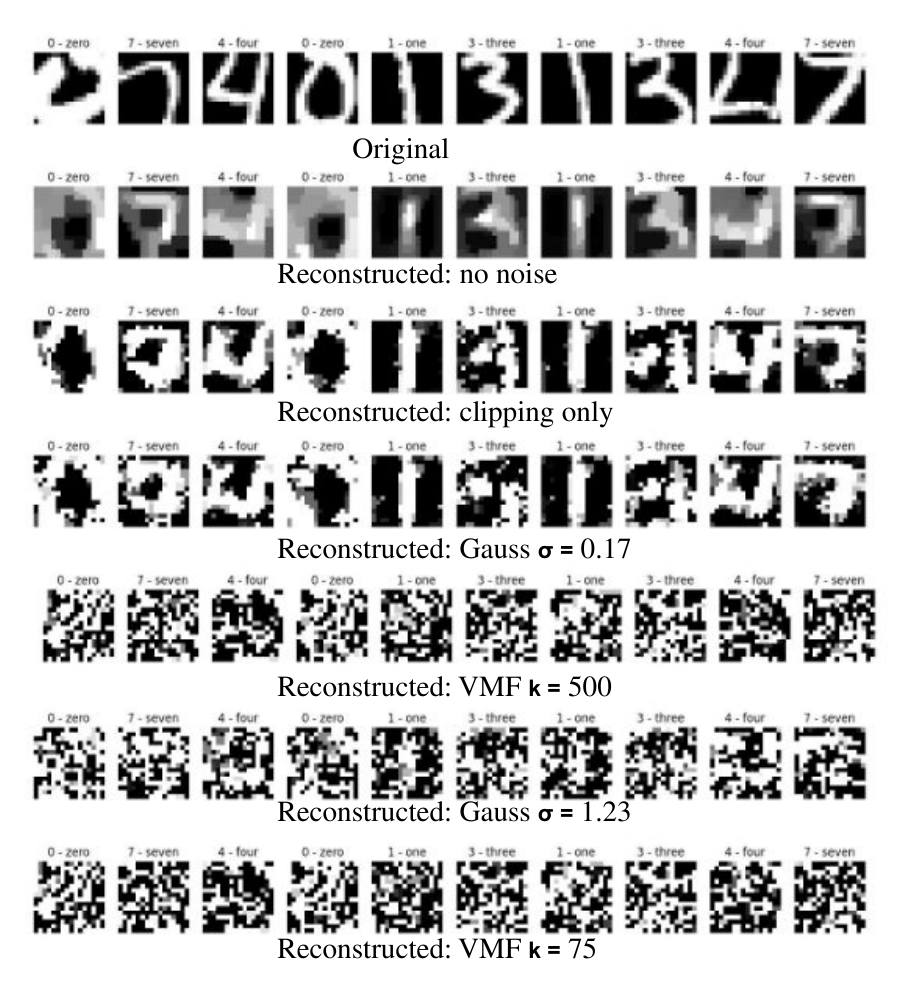}} &
        
        \subfloat[FashionMNIST reconstructions]{\includegraphics[width=0.4\textwidth]{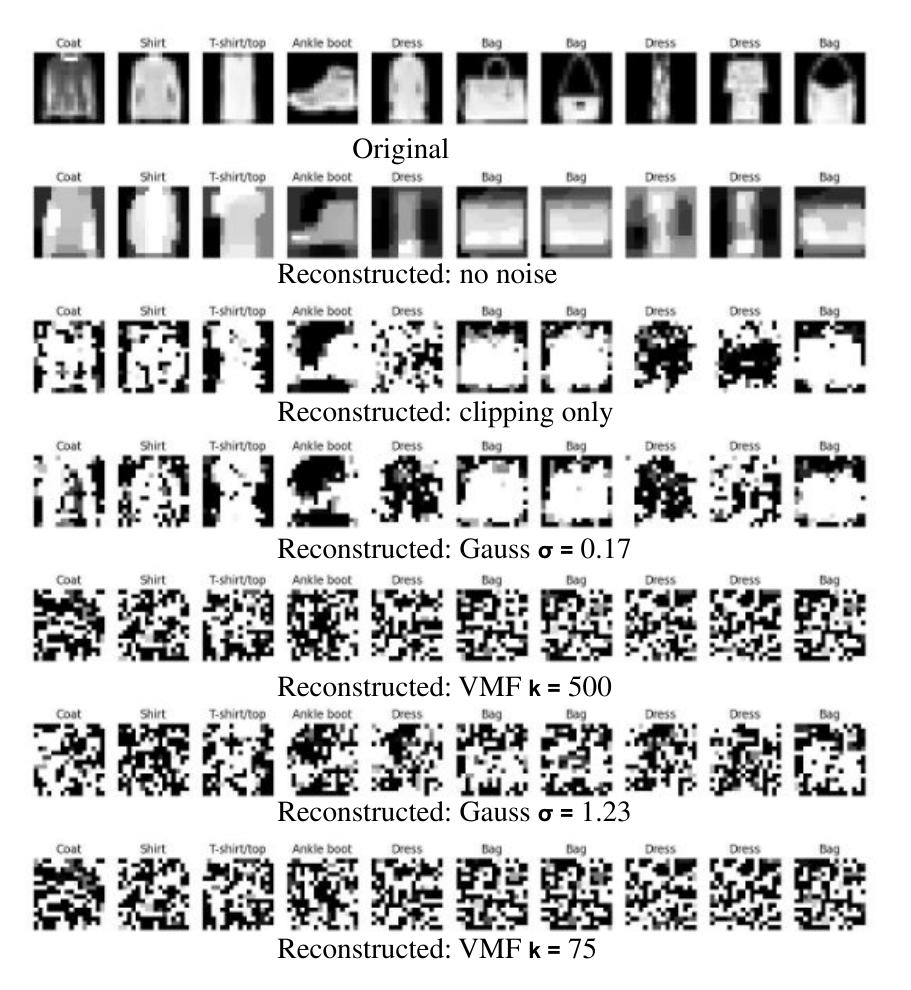}}
\end{tabular}
    \vspace{4mm}
    \caption{Reconstruction images for both datasets by the Inverting Gradients Attack (IGA).}\label{fig:reconstruction_images}
    \end{figure*}

We report in Figure~\ref{fig:reconstruction_plots} reconstruction measurements for both mechanisms under the same range of $\epsilon$ from Section \ref{sec:results-utility}.
For SSIM, bigger values mean better reconstruction; therefore, smaller values mean the mechanism protects better against reconstruction. For MSE, it is the opposite: smaller values mean the reconstructed images are closer to the original ones.

As with accuracy, there is no intersection point in terms of protection within each dataset. However, in contrast to the utility results, VMF always outperforms Gaussian in both datasets for this range of $\epsilon$. Gaussian protection deteriorates more than VMF for increasing $\epsilon$ for both metrics. SSIM scores are more stable across both datasets, with changes happening from the second decimal place. For MSE, we see a sharper deterioration in protection for Gaussian. We report individual points with their respective $\epsilon$, $\kappa$ and $\sigma$ in the Appendix.

We also show examples of images before and after reconstruction under different settings (Figure~\ref{fig:reconstruction_images}). We see that even for the setting in which a small DP noise is added images are barely recognisable for VMF. For Gaussian, however, stronger noise is needed (e.g., the last two rows, for which $\kappa = 500$ and $\sigma = 0.17$, which translates to $\epsilon=173$). 

We note that clipping alone is not enough to prevent image reconstruction from gradients by IGA: when the shared gradients are clipped by layer (i.e., the way Opacus handles clipping), numbers can be easily recognised in MNIST images, as the shape of some products in FashionMNIST images. We show specific reconstruction metric values in Table \ref{tab:reconstruction_metrics}.

\subsection{Bayes' capacity for reconstruction accuracy}
\label{sec:results-bayes}

\begin{figure}[ht!]
    {\includegraphics[width=0.45\textwidth]{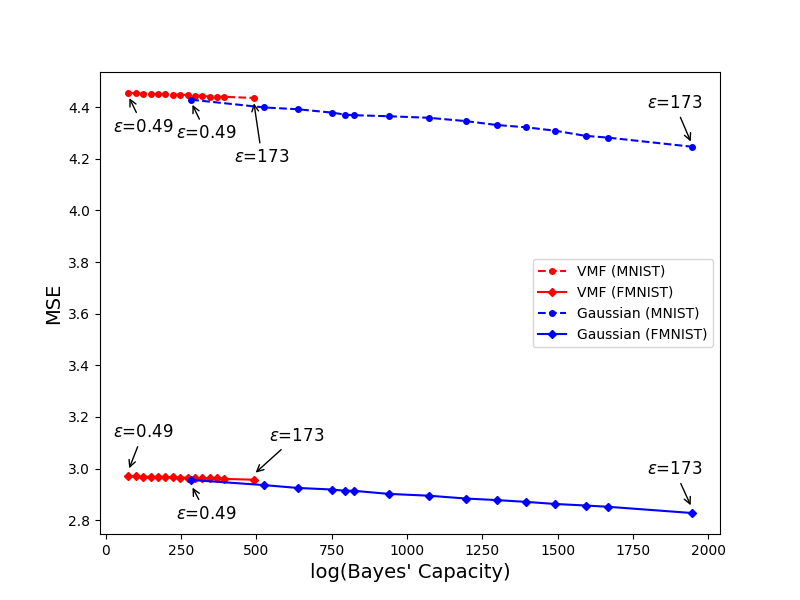}}
    \caption{Bayes' capacity is a better predictor for MSE than is $\epsilon$. In both datasets, the same $\epsilon$ values give different MSEs but similar Bayes' capacities give similar MSEs.}\label{fig:capacity}
\end{figure}

\begin{figure}[ht!]
    {\includegraphics[width=0.45\textwidth]{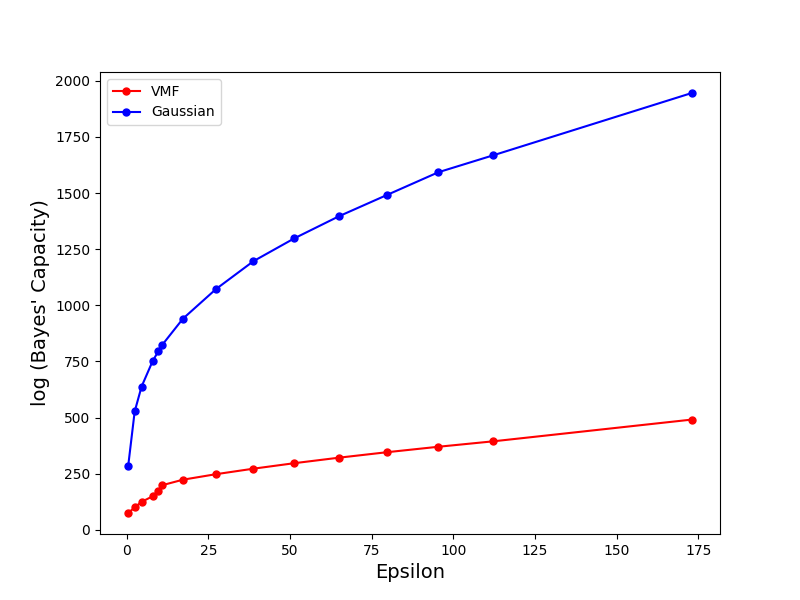}}
    \caption{Comparing leakage measures: $\epsilon$ vs Bayes' capacity for Gaussian and VMF mechanisms.}\label{fig:bayes_vs_epsilon}
\end{figure}

In \Sec{sec:comparing-bayes} we showed that a theoretical comparison of mechanisms (with regard to reconstruction attacks) can be made by comparing their Bayes' capacities. However, in the experimental literature on reconstruction in  machine learning (and in our experiments above), the MSE and SSIM measures are used as proxies for reconstruction accuracy. In this section, we dissect how well the Bayes' capacity measure performs with respect to the empirical measure of MSE, and compare this with the performance of DP.

The results shown in Figure~\ref{fig:reconstruction_plots} indicate that $\epsilon$ as a measure of privacy is not a good predictor for MSE. For the Gaussian mechanism, as $\epsilon$ increases (privacy decreases) MSE also decreases (reconstruction accuracy increases). While this might be expected, the same is not true for the VMF mechanism, for which MSE remains stable with increasing value of $\epsilon$. This suggests that comparing privacy mechanisms on the basis of their $\epsilon$-values is not a reliable approach for estimating the protections provided against reconstruction threats.

We next computed the Bayes' capacity for both Gaussian and VMF mechanisms and compared these with the corresponding MSE values. 
The results are plotted in Figure~\ref{fig:capacity}. We found that smaller values for the Bayes' capacity (better reconstruction protection) also correspond to larger MSE values (lower reconstruction accuracy), \emph{regardless of the mechanism used}. The plot identifies some values for $\epsilon$, again highlighting that $\epsilon$ in DP fails to capture MSE values across different mechanisms on the same dataset. 
These results indicate that our theoretical approach aligns well with the metrics used in practice to measure reconstruction attack success. Moreover, these results 
support the theory that Bayes' capacity is more useful than DP as a measure of capturing the vulnerability against reconstruction attacks.

Finally, to justify this theory, we plotted the Bayes' capacity versus values of $\epsilon$ for both mechanisms. 
The results are shown in Figure~\ref{fig:bayes_vs_epsilon}. We find that the $\epsilon$ parameter of DP and Bayes' capacities are in general \emph{incomparable}. That is, if $\epsilon_1 < \epsilon_2$ for different mechanisms, then we cannot draw conclusions about the Bayes' capacities for these mechanisms (and vice versa). What this means is that neither of these measures is completely robust against all attacks when used alone. 
This suggests that when applying privacy in practice, the goals of the adversary should be kept in mind when making decisions about which privacy measure to use. 

We remark that capacities (as modelled in the QIF framework) are not robust measures for comparing systems for all kinds of attacks. In fact, they represent an upper bound on all kinds of attacks, but, given a specific attack, one certain mechanism can provide better protection than another one, while, under a different attack, the order may be reversed. Our use of Bayes' capacity is justified by the fact that reconstruction attacks in DP-SGD correspond precisely to the leakage measured by capacity (i.e., one-try attack under a uniform prior). \emph{Refinement relations} are a much more robust method for comparing mechanisms~\cite{Chatzi:2019}, however these are usually too strong to hold in most settings. Our recommendation is to compare mechanisms using appropriate leakage measures when refinement doesn't hold. 

Our recommendation from these results is that while the DP parameter is useful as a measure against privacy threats in general, when comparing mechanisms for reconstruction attacks it is not safe to simply compare the values of $\epsilon$; one should use Bayes' capacity as an additional measure to choose a mechanism for a desired $\epsilon$ for DP.

\section{Conclusions}
\label{sec:conclusions}

In this paper, we investigated the question of how to compare the privacy protection provided by different formal privacy definitions in the context of ML. We focused on the VMF mechanism, which satisfies  metric privacy, and the Gaussian mechanism, which satisfies $(\epsilon, \delta)$-DP.
As privacy notion, we considered their ability to respond to the threat scenario determined by the reconstruction of an underlying dataset. 

We proposed two methods of comparison and evaluated them experimentally against known reconstruction attacks. 
The first method is based on RDP, used  to extract  $(\epsilon, \delta)$-DP guarantees for the VMF mechanism, to compare to those of the Gaussian mechanism,  in the context of DP-SGD. The second method is based on Bayes' capacity which measures the information leakage that can be exploited in a reconstruction attack.

Our empirical evaluation shows that, when we tune the parameters of the VMF mechanism to be equivalent to that of the  Gaussian in terms of RDP, the latter provides better utility, while VMF  consistently provides greater protection in preventing image reconstruction. Moreover, when we empirically match the utilities for the VMF and the Gaussian, the protection provided by VMF is still greater. 

We found experimentally that Bayes' capacity does correlate with the risk of reconstruction independently of the mechanism, unlike the parameters of DP, for which the correlation is mechanism-dependent. This suggests that  Bayes' capacity is a better metric for measuring the threats of reconstruction attack than the parameters of DP. 

More generally, our approach, using either or both techniques, provides a pathway for the comparison of mechanisms that satisfy different notions of formal privacy guarantees. 


\bibliographystyle{IEEEtran}
\bibliography{csf_main}

\appendices

\section{Proofs omitted from \Sec{sec:model}}~\label{app:proofs}

\subsection{Proof of \Cref{th:VMF_RenyiDiv_properties}}\label{app_proof: VMF_RenyiDiv_properties}
$(i)$
   As $\alpha\geq 1$, to show $\tau_{\alpha}$ is an increasing function of $\kappa$, it suffices to show that $\ln\left(\frac{I_\nu((2\alpha-1)\kappa)}{I_\nu(\kappa)}\right)$ is an increasing function of $\kappa$ when $\kappa\geq 1$. 
   Recalling that 
   the modified Bessel function of first kind $I_{\nu}(z)$, for any $\nu,z\geq 0$, can be written as 
   $$I_{\nu}(z)=\left(\frac{z}{2}\right)^{\nu}\sum\limits_{s=0}^{\infty}\frac{\left(\frac{z}{2}\right)^{2s}}{s!\Gamma\left(\nu+s+1\right)},$$
   where $\Gamma(.)$ is the Gamma function, and by setting $J=(2\alpha-1)\kappa$ and $J'=(2\alpha-1)\kappa'$, we essentially need to show that for every $\kappa\geq \kappa'\geq 1$ and a fixed $\nu\geq 0$ (i.e., for a fixed number of parameters $p$):
   \begin{align}
   &\ln\left(\frac{I_\nu((2\alpha-1)\kappa)}{I_\nu(\kappa)}\right)\geq \ln\left(\frac{I_\nu((2\alpha-1)\kappa')}{I_\nu(\kappa')}\right)\nonumber\\
   \iff&\frac{\left(\frac{J}{2}\right)^{\nu}\sum\limits_{s=0}^{\infty}\frac{\left(\frac{J}{2}\right)^{2s}}{s!\Gamma\left(\nu+s+1\right)}}{\left(\frac{\kappa}{2}\right)^{\nu}\sum\limits_{s=0}^{\infty}\frac{\left(\frac{\kappa}{2}\right)^{2s}}{s!\Gamma\left(\nu+s+1\right)}}\geq \frac{\left(\frac{J'}{2}\right)^{\nu}\sum\limits_{s=0}^{\infty}\frac{\left(\frac{J'}{2}\right)^{2s}}{s!\Gamma\left(\nu+s+1\right)}}{\left(\frac{\kappa'}{2}\right)^{\nu}\sum\limits_{s=0}^{\infty}\frac{\left(\frac{\kappa'}{2}\right)^{2s}}{s!\Gamma\left(\nu+s+1\right)}}\nonumber\\
\end{align}
As
\textbf{$\left(\frac{J\kappa'}{4}\right)^{\nu}=\left(\frac{J'\kappa}{4}\right)^{\nu}$}, it suffices to show:
\begin{align}
    &\lim_{N\rightarrow \infty}\sum\limits_{s=0}^{N}\frac{\left(\frac{J}{2}\right)^{2s}}{s!\Gamma\left(\nu+s+1\right)}\lim_{N\rightarrow \infty}\sum\limits_{s=0}^{N}\frac{\left(\frac{\kappa'}{2}\right)^{2s}}{s!\Gamma\left(\nu+s+1\right)})\nonumber\\
    \geq&\lim_{N\rightarrow \infty}\sum\limits_{s=0}^{N}\frac{\left(\frac{J'}{2}\right)^{2s}}{s!\Gamma\left(\nu+s+1\right)}\lim_{N\rightarrow \infty}\sum\limits_{s=0}^{N}\frac{\left(\frac{\kappa}{2}\right)^{2s}}{s!\Gamma\left(\nu+s+1\right)}\label{eq:VMF_RenyiDiv_kappa_inc2}
\end{align}
For any $\nu\geq 0$ and $s\in\mathbb{Z}_{\geq 0}$, let $F(\nu,s)=4^ss!\Gamma\left(\nu+s+1\right)$. Hence, for \eqref{eq:VMF_RenyiDiv_kappa_inc2}, we essentially need that for every $N\in\mathbb{Z}_{\geq 0}$:
\begin{align}
    &\sum\limits_{s=0}^{N}\frac{\left(J\right)^{2s}}{F(\nu,s)}\sum\limits_{s=0}^{N}\frac{\left(\kappa'\right)^{2s}}{F(\nu,s)}\geq\sum\limits_{s=0}^{N}\frac{\left(J'\right)^{2s}}{F(\nu,s)}\sum\limits_{s=0}^{N}\frac{\left(\kappa\right)^{2s}}{F(\nu,s)}\nonumber\\
    \iff&\sum\limits_{s=0}^{N}\frac{\left(J\kappa'\right)^{2s}}{F(\nu,s)^2}+\sum\limits_{\substack{s_1,s_2=0\\s_1>s_2}}^N\frac{\left(J\right)^{2s_1}\left(\kappa'\right)^{2s_2}+\left(J\right)^{2s_2}\left(\kappa'\right)^{2s_1}}{F(\nu,s_1)F(\nu,s_2)}\nonumber\\
    \geq&\sum\limits_{s=0}^{N}\frac{\left(J'\kappa\right)^{2s}}{F(\nu,s)^2} +\sum\limits_{\substack{s_1,s_2=0\\s_1>s_2}}^N\frac{\left(J'\right)^{2s_1}\kappa^{2s_2}+\left(J'\right)^{2s_2}\kappa^{2s_1}}{F(\nu,s_1)F(\nu,s_2)}\nonumber\\
    \iff&\sum\limits_{\substack{s_1,s_2=0\\s_1>s_2}}^N\frac{\left(J\right)^{2s_1}\kappa'^{2s_2}+\left(J\right)^{2s_2}\kappa'^{2s_1}}{F(\nu,s_1)F(\nu,s_2)}\nonumber\\
    &\geq\sum\limits_{\substack{s_1,s_2=0\\s_1>s_2}}^{N}\frac{\left(J'\right)^{2s_1}\left(\kappa\right)^{2s_2}+\left(J'\right)^{2s_2}\left(\kappa\right)^{2s_1}}{F(\nu,s_1)F(\nu,s_2)}\label{eq:VMF_RenyiDiv_kappa_inc3}
\end{align}
Hence, it suffices to show for every $0\leq s_2<s_1\leq N$:
\begin{align}
    &\frac{\left(J\right)^{2s_1}\left(\kappa'\right)^{2s_2}+\left(J\right)^{2s_2}\left(\kappa'\right)^{2s_1}}{F(\nu,s_1)F(\nu,s_2)}\nonumber\\
    &\geq\frac{\left(J'\right)^{2s_1}\left(\kappa\right)^{2s_2}+\left(J'\right)^{2s_2}\left(\kappa\right)^{2s_1}}{F(\nu,s_1)F(\nu,s_2)}\nonumber\\
    \iff&(2\alpha-1)^{2s_1}\kappa^{2s_1}{\kappa'}^{2s_2}+(2\alpha-1)^{2s_2}\kappa^{2s_2}{\kappa'}^{2s_1}\nonumber\\
    &\geq (2\alpha-1)^{2s_1}{\kappa'}^{2s_1}{\kappa}^{2s_2}+(2\alpha-1)^{2s_2}{\kappa'}^{2s_2}{\kappa}^{2s_1}\nonumber\\
    \iff& \kappa^{2s_1}{\kappa'}^{2s_2}\geq \kappa^{2s_2}{\kappa'}^{2s_1} 
    \iff\kappa^{2s_1-2s_2}\geq {\kappa'}^{2s_1-2s_2}\label{eq:VMF_RenyiDiv_kappa_inc4}
\end{align}
As $\kappa\geq \kappa'\geq 1$, and $s_1>s_2$, \eqref{eq:VMF_RenyiDiv_kappa_inc4} trivially holds.

$(ii)$
    In order to establish that ${\tau}_{\alpha}$ is a decreasing function of $\nu$ when $\nu\in\mathbb{N}$, we set $J=(2\alpha-1)\kappa$ and show that for $\nu,\nu'\in\mathbb{N}$ s.t. $\nu\geq\nu'$ and for $\alpha>1$, 
    \begin{align}
        &\frac{\nu}{\alpha-1}\ln\frac{1}{2\alpha-1}+\frac{1}{\alpha-1}\ln\frac{I_\nu((2\alpha-1)\kappa)}{I_\nu(\kappa)}\nonumber\\
        &\leq 
        \frac{\nu'}{\alpha-1}\ln\frac{1}{2\alpha-1}+\frac{1}{\alpha-1}\ln\frac{I_{\nu'}((2\alpha-1)\kappa)}{I_{\nu'}(\kappa)}\nonumber\\
        \iff&(\nu-\nu')\ln{(2\alpha-1)}\geq \ln{\frac{I_{\nu}(J)I_{\nu'}(\kappa)}{I_{\nu'}(J)I_{\nu}(\kappa)}}\label{eq:VMF_RenyiDiv_nu_dec1}
    \end{align}

    
Using the expression of the modified Bessel function of first kind as given in the first part of the proof and the fact that $\alpha>1$, we essentially need to have: 


\begin{align}
    &\ln{\frac{\sum\limits_{s=0}^{\infty}\frac{\left(\frac{J}{2}\right)^{2s}}{s!\Gamma\left(\nu+s+1\right)}\sum\limits_{s=0}^{\infty}\frac{\left(\frac{\kappa}{2}\right)^{2s}}{s!\Gamma\left(\nu'+s+1\right)}}{\sum\limits_{s=0}^{\infty}\frac{\left(\frac{J}{2}\right)^{2s}}{s!\Gamma\left(\nu'+s+1\right)}\sum\limits_{s=0}^{\infty}\frac{\left(\frac{\kappa}{2}\right)^{2s}}{s!\Gamma\left(\nu+s+1\right)}}}<0\nonumber\\
    \iff&\lim_{N\rightarrow\infty}\sum\limits_{s=0}^{N}\frac{\left(\frac{J}{2}\right)^{2s}}{s!\Gamma\left(\nu+s+1\right)}\lim_{N\rightarrow\infty}\sum\limits_{s=0}^{N}\frac{\left(\frac{\kappa}{2}\right)^{2s}}{s!\Gamma\left(\nu'+s+1\right)}\nonumber\\
    \leq&\lim_{N\rightarrow\infty}\sum\limits_{s=0}^{N}\frac{\left(\frac{J}{2}\right)^{2s}}{s!\Gamma\left(\nu'+s+1\right)}\lim_{N\rightarrow\infty}\sum\limits_{s=0}^{N}\frac{\left(\frac{\kappa}{2}\right)^{2s}}{s!\Gamma\left(\nu+s+1\right)}.\nonumber\\
&\text{It suffices to show that for all $N\in\mathbb{Z}_{\geq 0}$:}\nonumber\\
    &\sum\limits_{s=0}^{N}\frac{\left(\frac{J}{2}\right)^{2s}}{s!\Gamma\left(\nu+s+1\right)}\sum\limits_{s=0}^{N}\frac{\left(\frac{\kappa}{2}\right)^{2s}}{s!\Gamma\left(\nu'+s+1\right)}\nonumber\\
    &\leq\sum\limits_{s=0}^{N}\frac{\left(\frac{J}{2}\right)^{2s}}{s!\Gamma\left(\nu'+s+1\right)}\sum\limits_{s=0}^{N}\frac{\left(\frac{\kappa}{2}\right)^{2s}}{s!\Gamma\left(\nu+s+1\right)}\nonumber\\
    \iff&\sum\limits_{\substack{s_1,s_2=0\\s_1>s_2}}^N\left(\frac{\left(J\right)^{2s_1}\left(\kappa\right)^{2s_2}}{4^{s_1+s_2}s_1!s_2!\Gamma\left(\nu+s_1+1\right)\Gamma\left(\nu'+s_2+1\right)}\right.\nonumber\\
    &\left.+\frac{\left(J\right)^{2s_2}\left(\kappa\right)^{2s_1}}{4^{s_1+s_2}s_1!s_2!\Gamma\left(\nu+s_2+1\right)\Gamma\left(\nu'+s_1+1\right)}\right)\nonumber\\
    \leq&\sum\limits_{\substack{s_1,s_2=0\\s_1>s_2}}^N\left(\frac{\left(J\right)^{2s_1}\left(\kappa\right)^{2s_2}}{4^{s_1+s_2}s_1!s_2!\Gamma\left(\nu'+s_1+1\right)\Gamma\left(\nu+s_2+1\right)}\right.\nonumber\\
    &\left.+\frac{\left(J\right)^{2s_2}\left(\kappa\right)^{2s_1}}{4
    ^{s_1+s_2}s_1!s_2!\Gamma\left(\nu'+s_2+1\right)\Gamma\left(\nu+s_1+1\right)}\right)\label{eq:VMF_RenyiDiv_nu_dec4}
\end{align}
For $0\leq s_2<s_1\leq N$, setting $$H_1(s_1,s_2)=\frac{\left(\kappa\right)^{2s_1+2s_2}}{4^{s_1+s_2}s_1!s_2!\Gamma\left(\nu+s_1+1\right)\Gamma\left(\nu'+s_2+1\right)}\,\&$$ $$H_2(s_1,s_2)=\frac{\left(\kappa\right)^{2s_1+2s_2}}{4^{s_1+s_2}s_1!s_2!\Gamma\left(\nu+s_2+1\right)\Gamma\left(\nu'+s_1+1\right)},$$

showing \eqref{eq:VMF_RenyiDiv_nu_dec4} boils down to proving:

\begin{align}
    &\sum\limits_{\substack{s_1,s_2=0,\ldots,N\\s_1>s_2}}(2\alpha-1)^{2s_1}(H_2(s_1,s_2)-H_1(s_1,s_2))\nonumber\\
    \geq&\sum\limits_{\substack{s_1,s_2=0,\ldots,N\\s_1>s_2}}(2\alpha-1)^{2s_2}(H_2(s_1,s_2)-H_1(s_1,s_2))\nonumber
\end{align}
For any $s_1>s_2$, $(2\alpha-1)^{2s_1}-(2\alpha-1)^{2s_2}>0$. Thus, to conclude, it suffices to prove that for any $s_1>s_2$:
\begin{align}
    &H_2(s_1,s_2)\geq H_1(s_1,s_2)\nonumber\\
     \iff&\frac{\Gamma\left(\nu+s_2+T+1\right)}{\Gamma\left(\nu+s_2+1\right)}\geq\frac{\Gamma\left(\nu'+s_2+T+1\right)}{\Gamma\left(\nu'+s_2+1\right)}\label{eq:VMF_RenyiDiv_nu_dec6}\\
     &\text{ [letting $s_1=s_2+T$ for some $T\in\mathbb{N}$]}\nonumber
\end{align}

To establish \eqref{eq:VMF_RenyiDiv_nu_dec6}, we now show that $\frac{\Gamma\left(x+Q\right)}{\Gamma\left(x\right)}$ is an increasing function of $x\in\mathbb{R}^{+}$ for any $Q\in\mathbb{N}$. As $\Gamma(x+1)=x\Gamma(x)$ $\forall\,x\in\mathbb{R}^+$, $\frac{\Gamma\left(x+Q\right)}{\Gamma\left(x\right)}$ reduces to $(x+Q-1)\ldots x$, which is an increasing function for $x\in\mathbb{R}^+$.
\qed

\subsection{Proof of \Cref{prop:vmf:rdp:multi}}~\label{app_proof:vmf:rdp:multi}
The proof follows from Proposition~\ref{prop:vmf:rdp} and the fact that the R\'enyi divergence of order $\alpha$ between independent multi-variate VMF distributions is given by
\[D_\alpha\left(f_{\underline{p},\underline{\kappa}}^V(\underline{x})\|f_{\underline{p},\underline{\kappa}}^V(\underline{z})\right) = \sum_{i=1}^R D_\alpha\left(f_{{p_i},\kappa_i}^V(x_i)\|f_{{p_i},\kappa_i}^V(z_i)\right).\]

\qed

\subsection{Proof of \Cref{prop:flattenisbest}}\label{app_proof:flattenisbest}
Using \Cref{th:VMF_RenyiDiv_properties} (ii) \Cref{prop:vmf:rdp:multi}, we know that for $\kappa_i = \kappa$ and $p_i < p$ each $i$-th component in the summation~\eqref{eq:tau:alpha:multi} would be greater than that for $\kappa_i = \kappa$ and $p_i = p$. The result follows immediately.
\qed

\section{Proofs omitted from \Sec{sec:reconstruction} and \Sec{sec:bayes}}\label{app:bayes_proofs}

\subsection{Proof of \Cref{lem:BC_prop_0}}
Pre-processing by a deterministic channel has the effect of swapping rows or repeating rows. Neither of these actions involves removing a row of a channel or modifying the maximum element in the column of a channel. (Note that when we define channels we exclude the possibility of all zero columns - thus $C$ cannot remove rows from $D$). Since $\CBayes$ sums the column maxes for each column, it is unaffected by the pre-processing by $C$. The result follows.
\qed

\subsection{Proof of \Cref{lem:BC_prop_1}}\label{app_proof:BC_prop_1}
We note that the Bayes' capacity is computed as the sum of the column maxes of the channel.
    The channel which leaks nothing can be represented as a single columns of 1's, which has Bayes' capacity of 1. The channel which leaks everything is the identity matrix (excluding columns of all 0s), which for $N$ rows has exactly $N$ columns containing 1s, hence a Bayes' capacity of $N$.
\qed

\subsection{Proof of \Cref{corr:size of secret set}}\label{app_proof:size of secret set}
Denote by $CX^-$ the channel consisting of rows in $X' \setminus X$. Then we have $\CBayes(CX')=\sum_y \max_x {CX'}_{x,y}$
\begin{align*}
    &= \sum_y \max \{ \max_x {CX}_{x,y},\max_x {CX^-}_{x, y} \} \\
    &\geq \sum_y \max_x CX_{x,y}
       = \CBayes(CX)
\end{align*}
\qed

\subsection{Proof of \Cref{lem:BC_prop_2}}\label{app_proof:BC_prop_2}
    Follows immediately from \cite[Ch 10, Thm 10.7]{Alvim20:Book} which says that
    $ \CBayes(C{\cdot}D) \leq \min\{ \CBayes(C), \CBayes(D)\}$
\qed

\subsection{Proof of \Cref{lem:reconstruction-risk}}

Immediate from \Cref{eqn:dpsgd-leak} and \Cref{lem:BC_prop_0}.\qed

\subsection{Sketch proof of \Cref{th:BC_Gaussian}}\label{app_proof:BC_Gaussian}

  We are given a domain of $p$-dimensional vectors $\mathcal{X}$ which are mapped by \Alg{alg:sgd} to output vectors in $\mathbb{R}^p$ after clipping and averaging. We remark that the effect of clipping is that, given an unclipped $x \in \mathcal{X}$ and its clipped version $x_C \in \mathbb{B}_R^p$, the probability density function $f(x_C)$ centred at $x_C$ is exactly $f(x)$ for noise-adding mechanism $f$. (And the same holds true for averages, since averaging after clipping ensures a clipped average). Therefore we can write: 
\begin{align*}
    &\CBayes(G_{p, \sigma}) = \int_{\mathbb{R}^p} \sup_x G_{p, \sigma}(x)(y) ~dy \\
    &= \int_{y \in \mathbb{B}_R^{p}} \sup_x G_{p, \sigma}(x)(y) ~dy ~+~
     \int_{y \in \mathbb{R}^p \setminus \mathbb{B}_R^{p}} \sup_x G_{p, \sigma}(x)(y) ~dy
\end{align*}
For the first integral, the supremum for each $y$ occurs when $y = x$. Recalling the notation $f_{p,\sigma^2}^G(x)$ in \Def{def:gaussian} describing the probability density function for the Gaussian mechanism, we have:
\begin{align*}
    \int_{y \in \mathbb{B}_R^{p}} \sup_x G_{p, \sigma}(x)(y) ~dy = \int_{y \in \mathbb{B}_R^{p}} f_{p,\sigma^2}^G(y)(y) ~dy \\
    = f_{p,\sigma^2}^G(u)(u) \int_{y \in \mathbb{B}_R^{p}} ~dy 
    = \frac{1}{\sqrt{2\pi\sigma^2}^p} V(\mathbb{B}_R^p)
\end{align*}
where $u$ is any vector in $\mathbb{B}_R^p$ and $V(S)$ is the volume of $S$.

For the second integral, we observe that the supremum at $y$ occurs at the point $x$ on the surface of $\mathbb{B}_R^{p}$ which minimises the distance between $x$ and $y$. 
This occurs when $x$ is on the ray from $y$ through the origin. i.e. $x = R \frac{y}{\|y\|_2}$. Thus we have: 

\begin{align*}
    \int_{y \in \mathbb{R}^p \setminus \mathbb{B}_R^{p}} &\sup_x G_{p, \sigma}(x)(y) ~dy = \int_{y \in \mathbb{R}^p \setminus \mathbb{B}_R^{p}} f_{p,\sigma^2}^G(R \frac{y}{\|y\|_2})(y) dy \\
    &= \int_{\mathbb{R}^p \setminus \mathbb{B}_R^{p}} \frac{1}{\sqrt{2\pi\sigma^2}^p} e^{\frac{-\|y - R \frac{y}{\|y\|_2}\|_2^2}{2 \sigma^2}} dy \\
    &= \int_{\mathbb{R}^p \setminus \mathbb{B}_R^{p}} \frac{1}{\sqrt{2\pi\sigma^2}^p} e^{\frac{-(\|y\|_2 - R)^2}{2 \sigma^2}} dy\\
    &= \frac{1}{\sqrt{2\pi\sigma^2}^p} A(\mathbb{S}^{p-1}) \int_{r=R}^{\infty} r^{p-1} e^{\frac{-(r-R)^2}{2 \sigma^2}} dr \\
    &= \frac{1}{\sqrt{2\pi\sigma^2}^p} A(\mathbb{S}^{p-1}) \int_{s=0}^{\infty} (s+R)^{p-1} e^{\frac{-s^2}{2 \sigma^2}} ds 
\end{align*}
where $A(S)$ denotes the area of $(S)$ and $R$ is the radius of the ball $\mathbb{B}_R^p$. Note that the second-last line follows from change of variables to spherical coordinates.

Now, expanding $(s+R)^{p-1}$ and using the identity $\int_0^\infty x^{k-1} e^{\frac{-x^2}{2 \sigma^2}} dx = \frac{\Gamma(\frac{k}{2})}{2} (2 \sigma^2) ^ {\frac{k}{2}}$, we compute: 
\begin{align*}
     \int_{s=0}^{\infty}\!\! (s+R)^{p-1} e^{\frac{-s^2}{2 \sigma^2}} ds 
    &= \frac{1}{2} \sum_{i=0}^{p-1} \Gamma\!\left(\frac{p\!-\!i}{2}\right)\!(\sqrt{2} \sigma)^{p-i}\! { p\!-\!1 \choose i } R^i
\end{align*}
The result follows.
\qed

\subsection{Sketch proof of \Cref{th:BC_VMF}}\label{app_proof:BC_VMF}

In \Alg{alg:sgd2} (line 10) the domain $\mathcal{X}$ of vectors is $p$-dimensional vectors in $\mathbb{R}^p$ which are subsequently scaled to unit length so that the output domain is restricted to vectors in $\mathbb{S}^{p-1}$. For each input $x \in \mathcal{X}$ the VMF noise mechanism attains its maximum value at the point $y = x / \|x\|_2$. Since $\mathbb{S}^{p-1} \subseteq \mathcal{X}$, from \Eqn{eqn:bcapacity} we can choose each pointwise supremum over $\mathcal{X}$ at $y = x$. Using the notation $f_{p,\kappa}^V(x)$ from \Eqn{eq:VMF} describing the VMF density function, we have:
\begin{align*}
    &\CBayes(V_{p, \kappa}) = \int_{\mathbb{S}^{p-1}} \sup_x V_{p, \kappa}(x)(y)dy \\
    &=\int_{y \in \mathbb{S}^{p-1}} f_{p,\kappa}^V(y)(y)~ dy 
    = \frac{1}{C_{p,\kappa}}e^{\kappa} A(\mathbb{S}^{p-1})
\end{align*}
where $u$ denotes an arbitrary point on $\mathbb{S}^{p-1}$ (noting that the value of the VMF at any $u$ is independent of the mean $u$ chosen) and $A(S)$ denotes the area of $S$.
The result follows.
\qed

\section{Full results}\label{app:full_results}




\begin{table}[th!]
        \centering
        \begin{tabular}{cc|cc|cc}
        \hline
        \textbf{Mechanism} &
        \textbf{\makecell{$\kappa$ (VMF)/ \\ $\sigma$ (Gauss)}    } &
        \multicolumn{2}{c}{\textbf{MNIST}} & \multicolumn{2}{c}{\textbf{FMNIST}} \\

        &&\textbf{SSIM}  & \textbf{MSE} & \textbf{SSIM}  & \textbf{MSE}  \\ \hline

        None&-&.4543 & .1718 & .5099 & .1196\\ 
        Gauss&.0174& .0073&.2931 & .0257&.2331 \\ 
        VMF&500& .0114& .2904&.0300 & .2328\\ 
          
          \hline
         \end{tabular}
         \vspace{4mm}
        \caption{DLG reconstruction metrics for selected noises} 
        \label{tab:reconstruction_metrics_dlg}
    \end{table}

{
\setlength{\tabcolsep}{3pt}
\begin{table}[]
    \centering
    \tiny
    \begin{tabular}{c|ccccccccccccccc}
    \hline
        $\epsilon$ &0.49&2.48&4.59&7.97&9.72&10.9&17.25&27.38&38.84&41.02&64.98&79.68&95.44&112.28&173   \\
        $\kappa$ & 75 & 100 &125&150&175&200&225&250&275&300&325&350&375&400&500\\
        $\sigma$ &1.23&.660&.544&.461&.435&.420&.367&.321&.287&.282&.245&.229&.214&.204&.174\\ \hline
    \end{tabular}
    \caption{Mappings from $\epsilon$ to $\kappa$ and $\sigma$ used in our experiments.}
    \label{tab:eps_to_kappa_sigma}
\end{table}
}

\begin{figure}[h!]
    \centering
    \tiny
    \begin{tabular}{cc}
               
        \subfloat[MNIST reconstructions]{\includegraphics[width=0.15\textwidth]{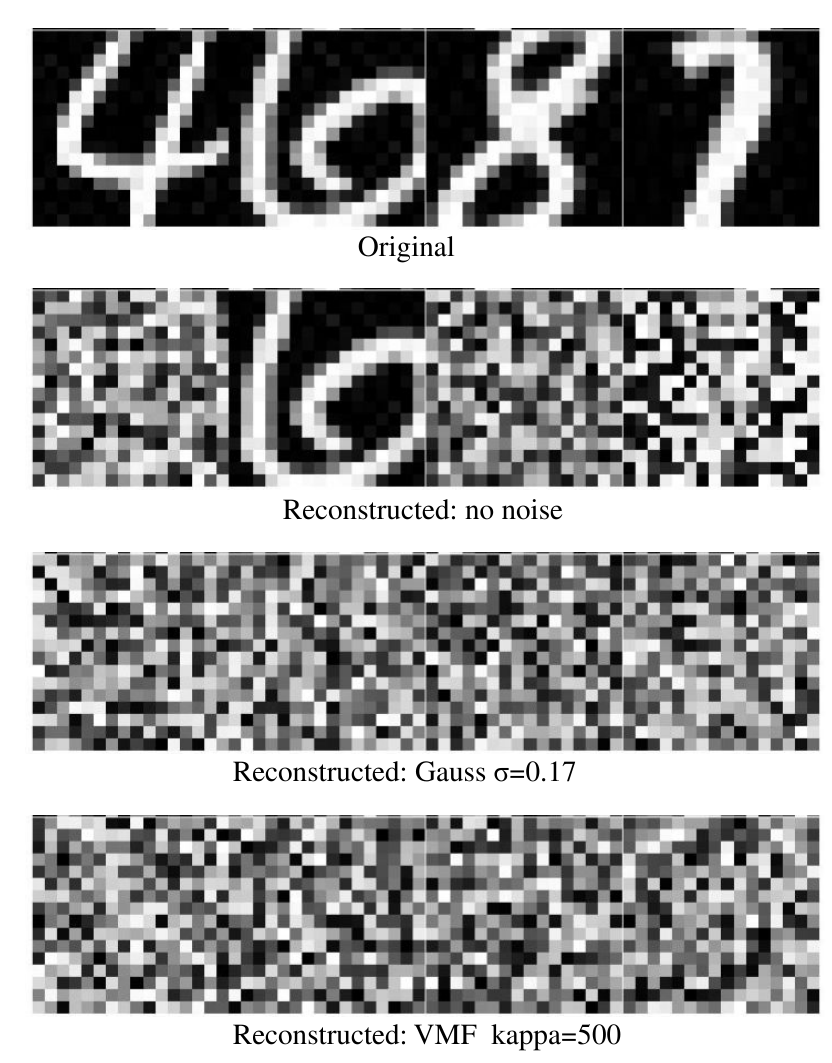}} &
        
        \subfloat[FashionMNIST reconstructions]{\includegraphics[width=0.15\textwidth]{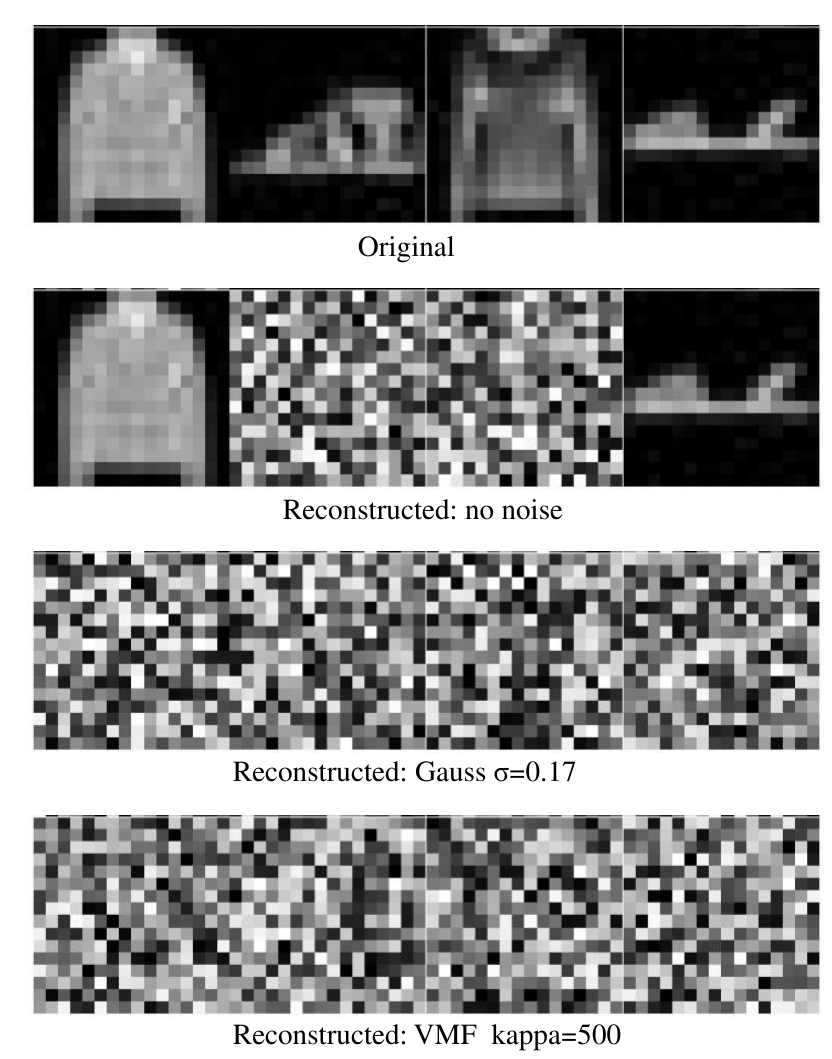}}
\end{tabular}
    \vspace{4mm}
    \caption{Reconstruction images for both datasets by the Deep Leakage from Gradients Attack (DLG).}\label{fig:reconstruction_images_dlg}
    \end{figure}

\subsection{Experiments with different batch sizes}

{
\setlength{\tabcolsep}{3.5pt}
\begin{table}[h!]
    \centering
    \tiny
    \begin{tabular}{cccccc|ccccc}
    \hline
    \textbf{Dataset} &
    \textbf{$\kappa$} &
    \textbf{Acc.\textsubscript{16}}  & \textbf{Acc.\textsubscript{32}} & \textbf{Acc.\textsubscript{64}}& \textbf{Acc.\textsubscript{128}} &
    \textbf{$\sigma$} &
    \textbf{Acc.\textsubscript{16}}  & \textbf{Acc.\textsubscript{32}} & \textbf{Acc.\textsubscript{64}}& \textbf{Acc.\textsubscript{128}} \\ \hline
     \multirow{9}{*}{MNIST}&75& .1815 &.1371 & .2736 &.3297& 1.23& .6848 &  .8073 &.8458&.8656\\
       &100& .2859 & .2424  & .3388 &.4182&.660 & .8112 & .8473  &.8729  & .8880\\          
       &150& .3405 &  .3598 & .4884 &.5293&.461& .8372 & .8653  & .8866  &.8977\\          
       &200& .4395 & .5663  & .5606 &.5948&.420& .8436 & .8698  &.8899 &.9000\\          
       &250 & .4767 & .5518  &.6229  &.6502&.321 & .8524 &  .8809 &.8958&.9034\\           
       &300& .6007 & .6625  & .6632 &.6940& .282& .8552 & .8870  & .8999&.9091\\           
       &350& .6373 & .7057  &.7061  &.7267& .229& .8618 & .8933  &.9031 &.9135\\         
       &400 & .6685 & .7543  &  .7569&.7535&.204 & .8686 & .8968  & .9057 &.9159\\  \hline       

       \multirow{9}{*}{\makecell{Fashion \\ MNIST}} &75&  .2727&  .2104 & .3821 &.4495&1.23& .5722 &  .6974 &  .7405 &.7489\\
       &100&.2917 & .3326  & .4146 &.4365&.660 & .7147 & .7327  & .7644 &.7879\\          
       &150& .4478 & .4453  & .4446 &.4592&.461& .7424 & .7591 &  .7850 &.7980\\          
       &200& .4622 & .5173  & .4997 &.4967&.420 & .7478 & .7670  & .7889  &.8025\\         
       &250& .4789 & .5051  & .5221 &.5227&.321 &.7611  & .7804  &  .7952 &8089\\          
       &300& .5284 & .5359  & .5484 &.5487& .282& .7653 & .7848  & .7983 &.8139\\           
       &350& .5421 & .5543  & .5925 &.5756& .229& .7727 & .7914  & .8039 &.8189 \\           
       &400& .5607 &  .5819 & .6586 &.6003&.204 & .7735 & .7962  & .8063 & .8204\\ \hline         
      
      \hline
     \end{tabular}
     \vspace{4mm}
    \caption{Accuracy values for varying batch sizes (subscript) for VMF (left) and Gaussian (right) mechanisms.} 
    \label{tab:utility_varying_batch}
\end{table}
 }
 
{
\setlength{\tabcolsep}{2pt}
\begin{table}[h!]
    \centering
    \tiny
    \begin{tabular}{ccccccccc|ccccccccc}
    \hline
    &\multicolumn{2}{c}{\textbf{Batch=16}}&\multicolumn{2}{c}{\textbf{Batch=32}}&\multicolumn{2}{c}{\textbf{Batch=64}} &\multicolumn{2}{c}{\textbf{Batch=128}}  &&\multicolumn{2}{c}{\textbf{Batch=16}}&\multicolumn{2}{c}{\textbf{Batch=32}}&\multicolumn{2}{c}{\textbf{Batch=64}}&\multicolumn{2}{c}{\textbf{Batch=128}} \\
    \textbf{$\kappa$} &
    \textbf{SSIM}  & \textbf{MSE} & \textbf{SSIM} & \textbf{MSE} & \textbf{SSIM} & \textbf{MSE} &
    \textbf{SSIM}  & \textbf{MSE}&
    \textbf{$\sigma$} &
    \textbf{SSIM}  & \textbf{MSE} & \textbf{SSIM} & \textbf{MSE} & \textbf{SSIM} & \textbf{MSE} &
    \textbf{SSIM}  & \textbf{MSE}\\ \hline

    \multicolumn{12}{c}{MNIST} \\\hline

    75&.0119  &4.177 & .0099& 4.355&-.0017 &  4.429&-.0035&4.396& 1.23& .0892 & 3.850&.0996 &3.966 &.1038 &  3.894 &.1046&3.923 \\
    100& .0135 & 4.165&.0110 &4.351 & -.0004&  4.422&-.0020&4.390&.660 & .1352 &3.550 &.1536 &3.646 &.1501 & 3.587  &.1586&3.572 \\        
       150& .0163 &4.137 &.0136 &4.334 & .0026&4.401  &.0003&4.378&.461&.1747  & 3.293& .1887&3.430 & .1812&  3.340  &.1928&3.303\\          
       200&.0197  &4.109 &.0160 &4.320 &.0044 &4.388  &.0026&4.364&.420&.1806  & 3.246& .1910&3.385 & .1877&3.300   &.1984&3.261 \\         
       250 & .0231 &4.090 & .0178&4.303 & .0065& 4.376 &.0040&4.357&.321 & .2115 &3.008 &.2139 &3.202 &.2127 & 3.124 &.2132&3.124\\           
       300& .0251 &4.075 &.0222 &4.272 &.0083 & 4.363 &.0059&4.348& .282& .2150 & 2.964& .2291& 3.070&.2194 &3.049    &.2235&3.042\\           
       350& .0275 &4.056 &.0250 &4.240 &.0110 &4.345  &.0080&4.340& .229& .2447 & 2.749&.2498 & 2.928& .2366& 2.909   &.2306&2.954\\        
       400&  .0306&4.037 &.0278 &4.224 &.0129 & 4.321 &.0096&4.335&.204 & .2581 &2.660 & .2595& 2.864& .2414& 2.866  &.2383&2.891\\  \hline    

        \multicolumn{12}{c}{Fashion MNIST} \\\hline

       75& .0049 & 2.788& .0040&2.869 & .0008& 2.922 &.0024&2.946& 1.23& .0129 & 2.801& .0197&2.767 &.0299& 2.676&.0472&2.499\\
    100& .0052 & 2.784&.0045 &2.863 &.0011 & 2.918 &.0030&2.939&.660 & .0259 &2.610 &.0371 &2.595 &.0507 &  2.453    &.0736&2.243 \\        
       150& .0060 &2.778 &.0055 &2.852 &.0018 & 2.913 &.0033&2.934&.461& .0475 & 2.465&.0512 &2.473 & .0685&   2.330 &.0972&2.097\\          
       200& .0072 &2.774 &.0063 &2.841 & .0038& 2.904 &.0041&2.925&.420& .0523 &2.430 &.0559 &2.439 & .0708& 2.309   &.0996&2.068\\         
       250 & .0081 &2.772 &.0076 & 2.828& .0044& 2.895 &.0046&2.920&.321 &.0731  &2.279 &.0672 & 2.342& .0806&  2.202  &.1159&1.965\\           
       300& .0096 &2.765& .0082& 2.822& .0055&  2.889&.0047&2.915& .282& .0855 & 2.196&.0694 &2.309 &.0864 & 2.168   &.1189&1.925\\           
       350& .0098 & 2.764& .0091& 2.816&.0071 &2.875  &.0052&2.906& .229& .1068 &2.070 &.0842 &2.204 & .0927&  2.108  &.1258&1.877\\        
       400& .0107 &2.760& .0112& 2.802&.0073 & 2.868 &.0059&2.898&.204 & .1187 &1.979 &.0953 & 2.145&.0961 &  2.071  &.1312&1.856\\  \hline      

     \end{tabular}
     \vspace{4mm}
    \caption{Reconstruction metrics for varying batch sizes for VMF (left) and Gaussian (right) mechanisms.} 
    \label{tab:reconstruction_varying_batch}
    \end{table}
}

\end{document}